\documentclass{article}

\usepackage[nonatbib, final]{neurips_data_2022}
\usepackage[numbers]{natbib}

\usepackage[utf8]{inputenc} 
\usepackage[T1]{fontenc}    
\usepackage{hyperref}       
\usepackage{url}            

\usepackage{booktabs}       
\usepackage{amsfonts}       
\usepackage{nicefrac}       
\usepackage{microtype}      
\usepackage{xcolor}         
\usepackage{xspace}
\usepackage{enumitem}
\usepackage{graphicx}

\usepackage{amsmath}
\usepackage{cleveref}
\usepackage{caption}
\usepackage{subcaption}
\usepackage{multirow}
\usepackage{bbold}
\usepackage{makecell}
\usepackage{wrapfig}

\usepackage{minitoc}

\newcommand{\name}{SafeBench\xspace}
\newcommand{\Sone}{Straight Obstacle\xspace}
\newcommand{\Stwo}{Turning Obstacle\xspace}
\newcommand{\Sthree}{Lane Changing\xspace}
\newcommand{\Sfour}{Vehicle Passing\xspace}
\newcommand{\Sfive}{Red-light Running\xspace}
\newcommand{\Ssix}{Unprotected Left-turn\xspace}
\newcommand{\Sseven}{Right-turn\xspace}
\newcommand{\Seight}{Crossing Negotiation\xspace}
\newcommand{\collisionrate}{collision rate\xspace}
\newcommand{\runredlight}{frequency of running red lights\xspace}
\newcommand{\runstopsign}{frequency of running stop signs\xspace}
\newcommand{\outofroad}{average distance driven out of road\xspace}
\newcommand{\followroute}{route following stability\xspace}
\newcommand{\routecompletion}{average percentage of route completion\xspace}
\newcommand{\timespent}{average time spent to complete the route\xspace}
\newcommand{\acceleration}{average acceleration\xspace}
\newcommand{\yawvelocity}{average yaw velocity\xspace}
\newcommand{\laneinvasion}{frequency of lane invasion\xspace}
\newcommand{\overallscore}{overall score\xspace}
\newcommand{\selectionrate}{selection rate\xspace}
\newcommand{\collisionrateabbr}{CR\xspace}
\newcommand{\runredlightabbr}{RR\xspace}
\newcommand{\runstopsignabbr}{SS\xspace}
\newcommand{\outofroadabbr}{OR\xspace}
\newcommand{\followrouteabbr}{RF\xspace}
\newcommand{\routecompletionabbr}{Comp\xspace}
\newcommand{\timespentabbr}{TS\xspace}
\newcommand{\accelerationabbr}{ACC\xspace}
\newcommand{\yawvelocityabbr}{YV\xspace}
\newcommand{\laneinvasionabbr}{LI\xspace}
\newcommand{\overallscoreabbr}{OS\xspace}
\newcommand{\selectionrateabbr}{SR\xspace}
\newcommand{\rawscenarionumber}{$3,140$\xspace}
\newcommand{\finalscenarionumber}{$2,352$\xspace}
\newcommand{\Learningtocollide}{Learning-to-collide\xspace}
\newcommand{\AdvSim}{AdvSim\xspace}
\newcommand{\CarlaGenerator}{Carla Scenario Generator\xspace}
\newcommand{\AdvTraj}{Adversarial Trajectory Optimization\xspace}
\newcommand{\Learningtocollideabbr}{LC\xspace}
\newcommand{\AdvSimabbr}{AS\xspace}
\newcommand{\CarlaGeneratorabbr}{CS\xspace}
\newcommand{\AdvTrajabbr}{AT\xspace}

\newcommand{\m}[1]{{\color{black}#1}}

\title{\name: A Benchmarking Platform for Safety Evaluation of Autonomous Vehicles}

\author{%
  \thanks{Equal Contribution}$\,$ Chejian Xu$^1$, $^*$Wenhao Ding$^2$, Weijie Lyu$^1$, Zuxin Liu$^2$, \\
  \bf Shuai Wang$^2$, Yihan He$^2$, Hanjiang Hu$^2$, Ding Zhao$^2$, Bo Li$^1$ \\
  $^1$University of Illinois at Urbana-Champaign\ \ \ $^2$Carnegie Mellon University\\
  \texttt{\small\{chejian2,wlyu3,lbo\}@illinois.edu, dingzhao@cmu.edu}  \\
  \texttt{\small \{wenhaod,zuxinl,shuaiwa2,yihanhe,hanjianh\}@andrew.cmu.edu} \\ 
}

\begin{document}

\maketitle

\begin{abstract}

As shown by recent studies, machine intelligence-enabled systems are vulnerable to test cases resulting from either adversarial manipulation or natural distribution shifts. This has raised great concerns about deploying machine learning algorithms for real-world applications, especially in safety-critical domains such as autonomous driving (AD). 
On the other hand, traditional AD testing on naturalistic scenarios requires hundreds of millions of driving miles due to the high dimensionality and rareness of the safety-critical scenarios in the real world.
As a result, several approaches for autonomous driving evaluation have been explored, which are usually, however, based on different simulation platforms, types of safety-critical scenarios, scenario generation algorithms, and driving route variations.
Thus, despite a large amount of effort in autonomous driving testing, it is still challenging to compare and understand the effectiveness and efficiency of different testing scenario generation algorithms and testing mechanisms under similar conditions.
In this paper, we aim to provide the \textit{first} unified platform \name to integrate different types of safety-critical testing scenarios, scenario generation algorithms, and other variations such as driving routes and environments.
In particular, we consider $8$ safety-critical testing scenarios following National Highway Traffic Safety Administration (NHTSA) and develop $4$ scenario generation algorithms considering $10$ variations for each scenario.
Meanwhile, we implement $4$ deep reinforcement learning-based AD algorithms with $4$ types of input (e.g., bird’s-eye view, camera) to perform fair comparisons on \name. We find our generated testing scenarios are indeed more challenging and observe the trade-off between the performance of AD agents under benign and safety-critical testing scenarios. 
We believe our unified platform \name for large-scale and effective autonomous driving testing will motivate the development of new testing scenario generation and safe AD algorithms.
\name is available at \url{https://safebench.github.io}.
\end{abstract}

\section{Introduction}
\label{sec:intro}




Innovations driven by recent progress in machine learning (ML) have shown human-competitive performance in sensing \citep{Silver1140}, decision-making \citep{he2015delving}, and manipulation \citep{agostinelli2019solving}.
However, several studies 
have shown that when such powerful ML models are exposed to adversarial attacks they can be fooled, evaded, and misled in ways that would have profound security implications: image recognition, natural language processing, and audio recognition systems have all been attacked~\citep{xiang2019generating,huang2017adversarial,huang2011adversarial,yuan2018commandersong}. 
As ML-based models and approaches have expanded to real-world \textit{safety-critical applications}, such as Autonomous Driving (AD), the question of safety is becoming a crux for the transition from theories to practice \citep{national2017automated,matheny2019artificial}, and it is vitally important to quantitatively and efficiently evaluate the robustness or safety of safety-critical applications before their massive production and deployment.
As listed in the \textit{National Artificial Intelligence Research and Development Strategic Plan} \citep{national2019national}, developing effective evaluation methods for AI and ML is considered one of the top priorities. Failing to meet this demand will cause death, stifle innovations, and hurt our economy, among other socially responsible issues.

\underline{\textit{Challenges.}}
Despite the great importance of safety evaluation for AD algorithms, it is challenging to comprehensively and quantitatively evaluate AD algorithms due to both \textit{real-world data} and \textit{evaluation design} challenges.
First, in practice, the safety-critical driving scenarios are ``rare'' -- can be found by driving every $30,000$ miles~\citep{disengagement}, which leads to the fact that current AD testing requires driving millions of miles with large economic and environmental costs. 
In addition, such rarity also requires the evaluation methods to have an accelerated feature with a probabilistic convergence guarantee to avoid being over-optimistic. Previous work \citep{Zhao2016f,o2018scalable} solve this problem for abstract simple models by using large deviation theories such as importance sampling (IS) and cross entropy (CE) \citep{bucklew2013introduction}. However, these approaches are shown to have reached bottlenecks when dealing with ML algorithms with increasing complexity. In fact, recent studies \citep{arief2020deep} have shown that these classical IS/CE based approaches and tools may consistently underestimate the risk when dealing with complex systems. Moreover, such peril has been identified in different evaluation approaches \citep{zhao2016accelerated,zhao2018accelerated,huang2018versatile,huang2017sequential,huang2018rare,huang2018synthesis}, which have already been adopted by industry \citep{ubercollaboration} and test agencies \citep{mcityreport} in the U.S. to assess the safety of AVs. 
Second, although several learning-based scenario generation approaches are later proposed to overcome the above challenge~\citep{najm2007pre,wang2021advsim,calo2020generating,wen2020scenario}, existing evaluation tools and platforms are usually based on their own design, such as dataset selection, safety-critical scenario definition and generation, evaluation metrics, and input types.
This makes it very challenging to fairly compare different AD algorithms or interpret different evaluation results.

In this paper, we focus on designing and developing the \textit{first} unified \textbf{robustness and safety evaluation platform for AD algorithms, \name}. In particular, we design \name based on the open-sourced simulation platform Carla~\citep{dosovitskiy2017carla}. \name consists of $4$ modules, including \textit{Agent Node}, \textit{Ego Vehicle}, \textit{Scenario Node}, and \textit{Evaluation Node}.
Based on our platform, we systematically evaluate the AD algorithms on \finalscenarionumber generated safety-critical testing scenarios, such as \textit{\Sone} and \textit{\Sthree} together with other benign scenarios.
For each safety-critical \textit{scenario}, we implement $4$ scenario \textit{generation algorithms} for comparison. In addition, for each scenario, we select $10$ diverse \textit{driving routes} to ensure the generalization of our evaluation results. 
We report the evaluation results based on $10$ metrics, such as \textit{\collisionrate}, \textit{\runredlight}, and \textit{\routecompletion}.
Finally, we developed $4$ reinforcement learning-based AD algorithms with different perceptual capabilities on \name. Specifically, we provide $4$ input types, ranging from low-dimensional state representations to complicated visual inputs. 
Based on our comprehensive evaluation, we find that (1) there is a performance trade-off for different AD algorithms under benign and safety-critical scenarios, (2) some safety-critical scenarios have higher transferability across AD algorithms, (3) different scenario generation algorithms achieve different levels of effectiveness even when generating the same scenario,  
(4) different AD algorithms achieve advantages over others under different metrics. Our findings suggest that testing AD algorithms on high-quality safety-critical scenarios is  necessary and can largely improve testing efficiency, and we should consider a combination of testing scenarios and generation algorithms for effective testing.

\underline{\textit{Contributions.}} In this work, we aim to provide the first \textbf{unified} evaluation platform for different AD algorithms by generating diverse safety-critical scenarios with different generation algorithms and evaluation metrics. Our evaluation platform \name includes the following properties.
\begin{itemize}[leftmargin=*]
    \item \textbf{Unified benchmarking platform with modularized design}. Our evaluation platform consists of 4 modules, including \textit{Ego vehicle}, \textit{Agent}, \textit{Scenario}, and \textit{Evaluation}. It is also flexible to replace, add, or delete modules for future functionalities and evaluations.
    \item \textbf{Comprehensive coverage of safety-critical scenario generation}. In \name, we have integrated \finalscenarionumber testing scenarios, which have provided comprehensive coverage of known safety-critical scenarios in the real world, and it is flexible to add more testing scenarios by applying generation methods on new template scenarios.
    \item \textbf{Comprehensive coverage of scenario generation algorithms}. For each testing safety-critical scenario, we developed 4 generation algorithms, so that we are able to not only evaluate AD safety on the scenario level, but also on the generation algorithm level.
    \item \textbf{Diverse metrics on safety measurement of different AD algorithms}. We report our evaluation based on $10$ evaluation metrics, based on three levels: safety, functionality, and etiquette.
    \item \textbf{General leaderboard of safety evaluation and extensible findings}. We provide a comprehensive leaderboard for the robustness and safety evaluation of $4$ AD algorithms, and we observe different performances of these AD algorithms under different controllable settings.
    \item \textbf{High flexibility and effectiveness}. Our evaluation platform is flexible to be integrated into other simulation platforms and different devices. Once the AD algorithm is trained, it is very effective to be tested on our generated testing scenarios.
    
\end{itemize}

\begin{table}
    \caption{\m{Comparison of Evaluation Platforms}}
    \centering
        \small{
    \label{tab:platform_comparison}
    \setlength{\tabcolsep}{3.75pt}
    \begin{tabular}{c|c|c|c|c|c|c}
    \toprule
    \multirow{2}{*}{Simulator}   
    & \multicolumn{2}{c|}{\makecell[c]{Safety-critical Scenarios}} 
    & \multirow{2}{*}{\makecell[c]{Realistic \\ Perception}}
    & \multirow{2}{*}{\makecell[c]{Customized \\ Scenario}} 
    & \multirow{2}{*}{\makecell[c]{Backend}}
    & \multirow{2}{*}{\makecell[c]{Baselines}} \\
    & \scriptsize{\makecell{Adversary-based}} & \scriptsize{\makecell{Knowledge-based}} & & & \\
    \midrule
    \name                                    & \checkmark  & \checkmark  & \checkmark  & \checkmark  & CARLA  & \checkmark   \\
    \midrule
    Scenario Runner~\cite{scenariorunner}        & $\times$    & \checkmark  & \checkmark  & \checkmark  & CARLA  & $\times$    \\
    DI-Drive Casezoo~\cite{didrive}              & $\times$    & \checkmark  & \checkmark  & \checkmark  & CARLA  & \checkmark  \\
    SUMMIT~\cite{cai2020summit}                  & $\times$    & $\times$    & \checkmark  & $\times$  & UE4  & \checkmark  \\
    Scenario Studio~\cite{zhou2020smarts}        & $\times$    & $\times$    & $\times$  & \checkmark  & SMARTS  & \checkmark   \\
    CommonRoad~\cite{wang2021commonroad}         & $\times$    & $\times$    & \checkmark  & \checkmark  & None  & $\times$  \\
    CausalCity~\cite{mcduff2021causalcity}       & $\times$    & $\times$    & \checkmark  & \checkmark  & UE4  & $\times$  \\
    MetaDrive~\cite{li2021metadrive}             & $\times$    & $\times$    & \checkmark  & \checkmark  & Panda3D  & \checkmark  \\
    highway-env~\cite{highway-env}               & $\times$    & $\times$    & $\times$  & \checkmark  & None  & $\times$  \\
    SUMO NETEDIT~\cite{sumo_netedit}             & $\times$    & $\times$    & $\times$  & \checkmark  & SUMO  & $\times$   \\  
    SimMobilityST~\cite{azevedo2017simmobility}  & $\times$    & $\times$    & $\times$  & \checkmark  & None  & $\times$   \\
    L2R~\cite{herman2021learn}                   & $\times$    & $\times$    & \checkmark  & \checkmark  & UE4  & $\times$   \\
    AutoDRIVE~\cite{samak2021autodrive}          & $\times$    & $\times$    & \checkmark  & \checkmark  & Unity  & $\times$  \\
    Deepdrive~\cite{deepdrive}                   & $\times$    & $\times$    & \checkmark  & \checkmark  & UE4  & $\times$    \\
    esmini~\cite{esmini}                         & $\times$    & $\times$    & \checkmark  & \checkmark  & Unity  & $\times$   \\
    AutonoViSim~\cite{best2018autonovi}          & $\times$    & $\times$    & \checkmark  & \checkmark  & PhysX  & $\times$   \\
    \bottomrule
    \end{tabular}
        }
\end{table}

\section{Related work}

Existing AD algorithm evaluation approaches and platforms can be categorized into three types based on how the testing driving scenarios are generated. First, the \textbf{data-driven} based generation and testing approaches~\citep{scanlon2021waymo, knies2020data, ding2018new, ding2020cmts} focus on real-world data sampling and distribution density estimation. This line of research is able to  model the real-world driving conditions, while requiring a large number of data collection to capture the ``rare'' safety-critical scenarios for testing. Second, the \textbf{adversary-based} generation and testing approaches~\citep{ding2021multimodal, zhang2022adversarial, feng2021intelligent} model the surrounding agents (e.g., vehicles and pedestrians) as adversarial agents to generate safety-critical driving scenarios. Third, the \textbf{knowledge-based} generation and testing approaches~\citep{ding2021semantically, wang2021commonroad, bagschik2018ontology} aim to integrate domain knowledge such as traffic rules as additional constraints to guide the testing scenario generation process. Recently, the latter two categories have shown efficient and effective evaluation results under specific driving environments and settings, and therefore we mainly focus on them in this work. However, existing driving scenario generation and testing approaches are developed on different platforms with different AD algorithms and sensor configurations, etc., making it challenging to directly compare the effectiveness of different testing scenarios, scenario generation algorithms, and the safety of AD algorithms. 
Thus, in this work we will provide the \textit{first} unified platform \name, to generate safety-critical scenarios with different algorithms considering a range of environments and configurations for fair comparison based on a comprehensive set of evaluation metrics. 
\m{In addition, several works have been conducted to test the safety of autonomous vehicles from the software testing perspective~\cite{abdessalem2018testing}, which mainly focuses on identifying the safety violations from the software level. Such testing frameworks can be integrated into \name as well for comprehensive testing.} 


\paragraph{\m{Comparison with other AD evaluation platforms}}

\m{
To accurately posit our \name platform in the AD evaluation area, we summarize existing platforms developed for autonomous vehicle evaluation and compare them with our platform in~\Cref{tab:platform_comparison}. We notice that very few of them consider safety-critical scenarios and the number of scenarios in existing evaluation platforms is very limited.
}

\section{\name: benchmarking platform for safety evaluation}

In this section, we will first provide an overview of our platform \name, followed by the details of our developed scenario generation algorithms and variants, as well as the evaluation metrics.

    \subsection{Platform structure}
    
        \begin{figure}
    \centering
    \includegraphics[width=1.0\textwidth]{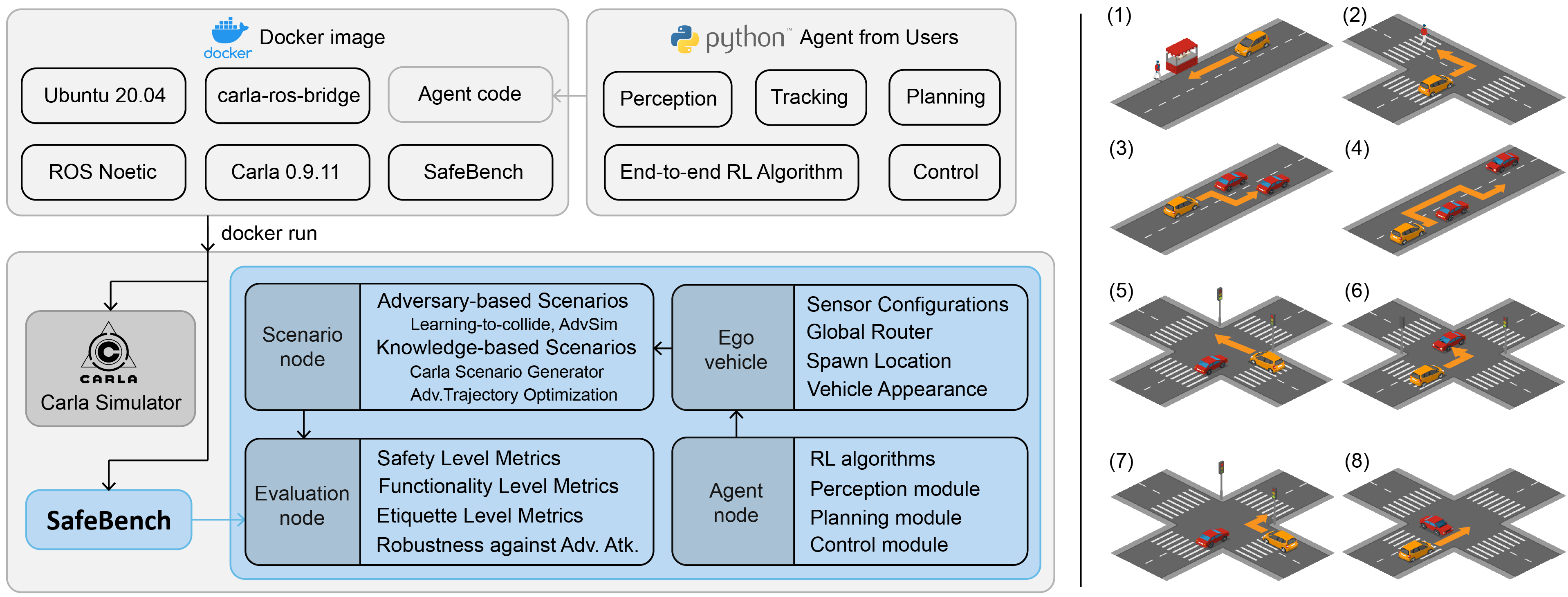}
    \caption{\small \textbf{Left:} Framework overview of \name. \textbf{Right:} 8 safety-critical driving scenarios - (1) \Sone (2) \Stwo (3) \Sthree (4) \Sfour (5) \Sfive (6) \Ssix (7) \Sseven (8) \Seight.}
    \label{fig:framework}
\end{figure}

        \textbf{Overview.} In~\Cref{fig:framework}, we show the structure of our \name platform. This platform runs in the Docker\m{~\citep{merkel2014docker}} container and is built upon the Carla simulator~\citep{dosovitskiy2017carla}. We use ROS\m{~\citep{ros}} for communication between the modules in the platform. In particular, \name consists of 4 components (nodes) as introduced in the following.
        
        \textbf{Ego vehicle} provides a virtual vehicle including the configurations of sensors (e.g., the positions and parameters of LiDAR, Camera, and Radar), the global planner, and the appearance of the vehicle. The testing AD algorithms are deployed in this node to interact with the driving scenarios. Users can change the configuration of this node to satisfy the requirement of their  algorithms.
        
        \textbf{Agent node} is designed to train and manage AD algorithms for  ego and surrounding vehicles, taking as input the observation information from the testing scenarios and outputting the controlling signals. AD algorithms managed by this node can  be trained on our platform.

        \textbf{Scenario node} is the core part of \name, which is responsible for organizing and generating  testing scenarios. These scenarios control the  behaviors of  traffic participants (e.g., pedestrians and surrounding vehicles) and static driving environments (e.g.,  road layout and status of traffic lights). 

        \textbf{Evaluation node} is designed to provide comprehensive evaluations  by testing different AD algorithms under diverse generated driving scenarios based on different metrics. The Evaluation Node collects all information during  testing and provides an evaluation summary on different  levels.

    \subsection{Safety-critical testing scenarios}
    \label{sec:scenario_definition}



        In this section, we first define the safety-critical traffic testing scenarios we considered in this work, containing 8 most representative and challenging driving scenarios of pre-crash traffic~\citep{najm2007pre} summarized by the National Highway Traffic Safety Administration (NHTSA).
        In addition, for each scenario, we design ten diverse driving routes that vary in terms of surrounding environments, number of lanes, road signs, etc.
        \m{Please see more detailed scenario definitions and route variants in \Cref{appendix:scenario_route}.}

        \paragraph{Pre-crash safety-critical scenarios.}
        We show the 8 pre-crash scenarios in the right part of~\Cref{fig:framework}. 
        In each scenario, the ego vehicle needs to drive along a pre-defined route and react to emergencies that occur on the road while driving. 
        Throughout the process, the ego vehicle should follow the traffic rules and avoid potential car accidents. 

        \begin{wrapfigure}{r}{0.340\textwidth}
     \centering
     \begin{subfigure}[t]{0.34\textwidth}
         \centering
         \includegraphics[width=\textwidth]{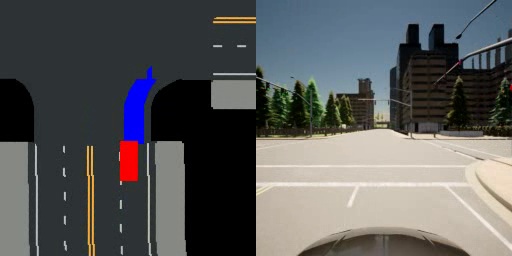}
     \end{subfigure}
     \begin{subfigure}[t]{0.34\textwidth}
         \centering
         \includegraphics[width=\textwidth]{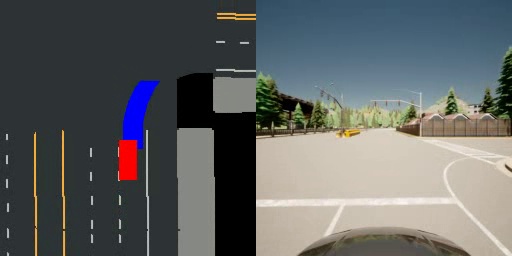}
     \end{subfigure}
        \caption{\small An example of route variants in \textit{\Stwo} scenario, consisting of a different number of lanes (2-lane vs. 3-lane road) and surrounding buildings.}
        \label{fig:variants}
    \vspace{-13mm}
\end{wrapfigure}
    
        \paragraph{Driving routes.}
        In practice, a driving scenario may involve many variants. For instance, small changes in the vehicle location or in the surrounding environment may lead to big changes in vehicle decision-making.
        In order to provide a more comprehensive safety evaluation, we design 10 driving routes for each safety-critical scenario.
        Each driving route has a sequence of pre-defined waypoints.
        Different driving routes of the same scenario may have a different number of lanes, different scenes (e.g., intersections, T-junctions, bridges, etc.), or different road signs, which restrict vehicle behaviors in different ways.
        We show $2$ example route variants of \textit{\Stwo} in \Cref{fig:variants}. 

    \subsection{Safety-critical scenario generation algorithms}
    \label{sec:scenario_generation}
    
        In this section, we detail how we collect and optimize safety-critical testing scenarios using different generation algorithms.
        Specifically, for each driving route mentioned above, we develop 4 algorithms to generate various testing samples.
        These algorithms mainly fall into two categories: adversary-based generation and knowledge-based generation.
        
        \subsubsection{Adversary-based generation}
            
            The state-of-the-art adversarial generation algorithms usually consist of two components: the scenario generator, and the victim model (i.e., the ego vehicle or tested AD agent). 
            Existing adversarial generation frameworks adopt different strategies to manipulate traffic scenarios, such as perturbing the position of surrounding vehicles (SVs) or forcing a cyclist to take an adversarial action, such that the victim model will crash into SVs and fail in the generated scenario. 
            To examine the safety and robustness of the tested AD agent against such adversarial scenarios, we select two representative algorithms as follows:
            ($i$) \textbf{\Learningtocollide (\Learningtocollideabbr)}~\citep{ding2020learning} is a black-box algorithm that optimizes the initial poses of a cyclist to attack the AD algorithm. 
            Following the default setting, we formulate the traffic scenarios as a series of auto-regressive building blocks and obtain the generated scenarios by sampling from the joint distribution of these blocks. 
            The policy gradient method REINFORCE~\citep{williams1992simple} is used to solve the scenario optimization problem. 
            In \Learningtocollideabbr, the authors only focus on generating \textit{\Stwo} scenario, so we adapt the method to all the $8$ scenarios and generate different initial conditions for all the driving routes.
            ($ii$) \textbf{\AdvSim (\AdvSimabbr)}~\citep{wang2021advsim} directly manipulates existing trajectories to perturb the driving paths of SVs, posing dangers to the tested AD agent.
            We follow the default setting and use the kinematic bicycle model~\citep{polack2017kinematic} to represent and calculate the full trajectory of SVs.
            Based on the results obtained by interacting with the driving environment, we optimize the trajectory parameters using the black-box search algorithm Bayesian Optimization~\citep{srinivas2010gaussian, ru2019bayesopt}.
            Similarly, in our experiments, we generate adversarial trajectories for all the route variants.

        \subsubsection{Knowledge-based generation}
            
            In the physical world, driving scenarios need to satisfy traffic rules and physical laws.
            Scenarios generated by adversarial algorithms, however, sometimes violate these rules.
            Therefore, we develop novel generation algorithms that integrate domain knowledge into the generation process.
            We select two representative algorithms as follows.
            ($i$) \textbf{\CarlaGenerator (\CarlaGeneratorabbr)}~\citep{scenariorunner} is a module built on the Carla Simulator~\citep{dosovitskiy2017carla} which uses rule-based methods to construct testing scenarios. 
            Following the standard process, we adopt the rules and use grid search to generate safety-critical scenario parameters for all the $8$ traffic scenarios.
            ($ii$) \textbf{\AdvTraj (\AdvTrajabbr)}~\citep{zhang2022adversarial} uses explicit knowledge as constraints to guide the scenario optimization process.
            We adopt the same constraints that needed to be satisfied and use the default PSO-based~\citep{poli2007particle} blackbox optimization for generating all kinds of testing scenarios in \name.
            
    \subsection{Evaluation metrics}
    \label{sec:evaluation_metric}
        
        In this section, we introduce the evaluation metrics used in \name.
        Specifically, we evaluate the performance of AD algorithms on 3 levels: \textit{Safety level}, \textit{Functionality level}, and \textit{Etiquette level}.
        Within each level, we design several metrics  focusing on different aspects.
        Finally, an \textit{\overallscore} is calculated as a weighted sum of all the evaluation metrics introduced below.
       
        \paragraph{Safety level}
        To evaluate the safety of given AD algorithms, we \m{follow existing works~\citep{carlachallenge, li2022metadrive} and} consider $4$ evaluation metrics focusing on serious violations of traffic rules: \textit{\collisionrate} (\collisionrateabbr), \textit{\runredlight} (\runredlightabbr), \textit{\runstopsign} (\runstopsignabbr), and \textit{\outofroad} (\outofroadabbr).
        Formally, we define the scenario trajectory as $\tau$, which is sampled from a scenario distribution $\mathcal{P}$, then the collisions that happened in one scenario after testing the AD algorithm can be represented as $c(\tau)$. Similarly, we obtain the number of running red lights $r(\tau)$, running stop signs $s(\tau)$, and distance driven out of road $d(\tau)$. The $4$ metrics are concretely calculated as: $\collisionrateabbr = \mathbb{E}_{\tau\sim\mathcal{P}}[c(\tau)]$, $\runredlightabbr = \mathbb{E}_{\tau\sim\mathcal{P}}[r(\tau)]$, $\runstopsignabbr = \mathbb{E}_{\tau\sim\mathcal{P}}[s(\tau)]$, and $\outofroadabbr = \mathbb{E}_{\tau\sim\mathcal{P}}[d(\tau)]$.
        
       
        \paragraph{Functionality level} 
        In each testing scenario, the AD agent is expected to follow and complete a specific route.
        This level of evaluation metrics is used to measure the functional ability of AD agents to finish such a task.
        \m{Inspired by previous works~\citep{carlachallenge, li2022metadrive},} we develop $3$ metrics as follows: \textit{\followroute} (\followrouteabbr), \textit{\routecompletion} (\routecompletionabbr), and \textit{\timespent} (\timespentabbr).
        To calculate \followrouteabbr, we use the average distance between the ego vehicle and the reference route during each testing $x(\tau)$. Then we calculate $\followrouteabbr = 1 - \mathbb{E}_{\tau\sim\mathcal{P}}[\min \left\{\frac{x(\tau)}{x_{max}}, 1\right\}]$,
        where $x_{max}$ is a constant indicating the maximum deviation distance.
        \routecompletionabbr is calculated as $\routecompletionabbr = \mathbb{E}_{\tau\sim\mathcal{P}}[p(\tau)]$,
        where $p(\tau)$ is the percentage of route completion of each testing scenario.
        \timespentabbr is the average time spent for completing the routes successfully: $\timespentabbr = \mathbb{E}_{\tau\sim\mathcal{P}}[t(\tau) | p(\tau) = 100\%]$,
        where $t(\tau)$ denotes the time cost of each testing scenario.
       
        \paragraph{Etiquette level}
        In practice, driver etiquette is an indicator of the driving skills of AD algorithms. 
        Here we \m{follow existing works~\citep{peng2020driving, huang2019developing} and} consider 3 metrics accordingly: \textit{\acceleration} (\accelerationabbr), \textit{\yawvelocity} (\yawvelocityabbr), and \textit{\laneinvasion} (\laneinvasionabbr).
        Similarly, these metrics are calculated as the expectation over all testing scenarios: $\accelerationabbr = \mathbb{E}_{\tau\sim\mathcal{P}}[acc(\tau)]$, $\yawvelocityabbr = \mathbb{E}_{\tau\sim\mathcal{P}}[y(\tau)]$, and $\laneinvasionabbr = \mathbb{E}_{\tau\sim\mathcal{P}}[l(\tau)]$,
        where $acc(\tau)$, $y(\tau)$, $l(\tau)$ denote the accelerations, yaw velocities, and number of lane invasions respectively.
        
        \paragraph{Overall score}
        To obtain an evaluation overview of the quality of AD algorithms, we aggregate all the metrics and report an \textit{\overallscore} (\overallscoreabbr), which is a weighted sum of the $10$ metrics introduced above.
        Specifically, the \overallscore is calculated as: $\overallscoreabbr = \sum_{i=1}^{10} w^i \times g(m^i)$,
        where $m^i$ is the $i^{th}$ metric, $w^i$ is the corresponding weight, 
        $g(m^i)$ is defined as 
        \begin{equation}
            g(m^i) = 
            \begin{cases}
            \frac{m^i}{m^i_{max}}, & m^i \text{is the higher the better} \\
            1-\frac{m^i}{m^i_{max}}, & m^i \text{is the lower the better} \\
            \end{cases}
        \end{equation}
        where $m^i_{max}$ is a constant indicating the maximum allowed value of $m^i$. More details of the constant\m{, parameters,} and weight selection are in \Cref{appendix:evaluation_metrics}.



\section{Benchmark evaluation on \name}
    In this section, we will first introduce the AD algorithms we will test which are based on different input state types, then illustrate our testing scenario generation and selection details, followed by our comprehensive benchmark results and corresponding observations and findings.

\subsection{AD algorithms tested on \name}

        We test various types of algorithms based on the safety-critical scenarios in \name.
        We particularly focus on reinforcement learning-based self-driving methods, since they require minimum domain knowledge of the overall system and driving scenarios~\cite{sallab2017deep, chen2019model,chen2021interpretable, kiran2021deep}. 
        One only needs to specify the reward function, action space, and state space, then train the agent by interacting with the scenario, and finally obtain a self-driving agent with reasonable performance.
        The reward function is given by a linear combination of the route following bonus, the collision penalty, the speeding penalty, and the energy consumption penalty.
        The action space is specified by the steering and throttle of the vehicle.
        
        We select $4$ representative deep RL methods for evaluation, including a stochastic on-policy algorithm -- Proximal Policy Optimization (PPO)~\cite{schulman2017proximal}, a stochastic off-policy method -- Soft Actor-Critic (SAC)~\cite{haarnoja2018soft}, and two deterministic off-policy approaches -- Deep Deterministic Policy Gradient (DDPG)~\cite{lillicrap2015continuous} and Twin Delayed DDPG (TD3)~\cite{fujimoto2018addressing}.
        To encourage the diversity of evaluation agents, we vary the state space to equip them with different perceptual capabilities.
        We design $4$ \textbf{state spaces} for each RL algorithm based on previous works~\cite{chen2019model,chen2021interpretable} as follows.
        The detailed model design and hyperparameters are presented in \cref{appendix:agent_implementation}.
        
        \begin{itemize}[leftmargin=*]
            \item \textbf{4D}. The basic observation type contains only $4$ dimensions of observation: distance to the waypoint, longitude speed, angular speed, and a front-vehicle detection signal. 
            \item \textbf{4D+Dir}. For a more complex observation type, we add another $7$ dimensions of observations, which are "Command (turn left, turn right or go straight)" and vectors that represent the direction of the ego vehicle, current waypoint, and target waypoint. 
            \item \textbf{4D+BEV}. We render the ego vehicle's local semantic map \m{using the information provided by CARLA} as the bird's-eye view (BEV) image, where the vehicles are represented by boxes. Lanes and routes are represented by line segments. 
            We incorporate the BEV image together with $4$ dimensional states to form this observation type.
            \item \textbf{4D+Cam}. This observation type includes an image captured by the front camera with 4D.
        \end{itemize}


\subsection{Driving scenarios for testing}

\textbf{Scenario generation.} 
We apply 4 safety-critical scenario generation algorithms to 8 template scenarios, each of which contains 10 diverse driving routes.
For each generation algorithm, we keep $9$ \m{or} $10$ testing scenarios based on their qualities.
Thus, in total, we generate \rawscenarionumber testing scenarios for evaluation.
We note that some scenario generation algorithms require a surrogate model to search for effective safety-critical configurations. 
For instance, we follow the setup of \Learningtocollideabbr~\citep{ding2020learning} to train a surrogate SAC model based on random benign scenarios. 

\begin{wrapfigure}{r}{0.4\textwidth}
    \centering
    \vspace{-20pt}
    \includegraphics[width=0.39\textwidth]{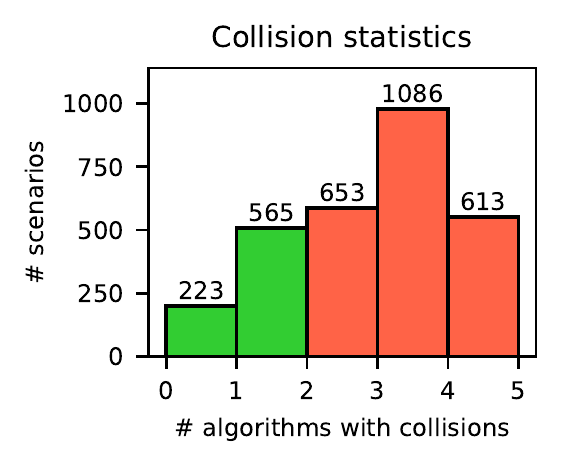}
    \vspace{-8pt}
    \caption{\small Collision statistics of generated scenarios before scenario selection. \textcolor{red}{Red} bars represent the selected ones with high collision rate. \textcolor{green}{Green} bars represent the unselected scenarios with low collision rate.}
    \vspace{-10pt}
    \label{fig:data_selection}
\end{wrapfigure}

\textbf{Scenario selection.} 
After collecting the raw testing scenarios, we select scenarios with desired properties.
Specifically, we test all the generated scenarios on $4$ AD algorithms with basic observation type and select scenarios that cause the most collisions. 
In \Cref{fig:data_selection}, we show a histogram of the  distribution for collisions.
We only keep scenarios that cause collisions for at least $2$ algorithms during testing, which is shown in red in \Cref{fig:data_selection}.
The selected testing scenarios have high transferability across AD algorithms and high risk levels, which further improves both the effectiveness and efficiency of AD evaluation.
After the selection, we obtain \finalscenarionumber testing scenarios in total. More details can be found in \Cref{appendix:scenario_statistics}.


\begin{table}[t]
\small
\centering
    \caption{\small \textbf{Statistics of scenario generation/selection}. We report \textit{\collisionrate} (\collisionrateabbr) before and after scenario selection (S-\collisionrateabbr) to measure the effectiveness of different scenario generation algorithms. The \textit{\overallscore} (\overallscoreabbr) before and after scenario selection (S-\overallscoreabbr) are used to demonstrate the safety-critical scenario generation capability of different algorithms. The \textit{\selectionrate} (\selectionrateabbr) is reported to evaluate the transferability of generation algorithms across AD agents. The last column shows the average over all the scenarios, with bold numbers indicating the best performance among the $4$ generation algorithms. \Learningtocollideabbr: \Learningtocollide, \AdvSimabbr: \AdvSim, \CarlaGeneratorabbr: \CarlaGenerator, \AdvTrajabbr: \AdvTraj, $\uparrow$/$\downarrow$: higher/lower the better.}
    \label{tab:generation_filtering}
{
{
\setlength{\tabcolsep}{3.75pt}
    \begin{tabular}{c|c|cccccccc|c}
    \toprule
        \multirow{3}{*}{\textbf{Metric}} & \multirow{3}{*}{\textbf{Algo.}} & \multicolumn{8}{c|}{\textbf{Traffic Scenarios}} & \multirow{3}{*}{\textbf{Avg.}} \\
        & & \scriptsize{\makecell{Straight \\ Obstacle}} & \scriptsize{\makecell{Turning \\ Obstacle}} & \scriptsize{\makecell{Lane \\ Changing}}  & \scriptsize{\makecell{Vehicle \\ Passing}} & \scriptsize{\makecell{Red-light \\ Running}} & \scriptsize{\makecell{Unprotected \\ Left-turn}} & \scriptsize{\makecell{Right-\\ turn}} & \scriptsize{\makecell{Crossing \\ Negotiation}} & \\
        \midrule
          \multirow{4}{*}{\collisionrateabbr $\uparrow$} & \Learningtocollideabbr & 0.320 & 0.140 & 0.560 & 0.920 & 0.410 & 0.630 & 0.458 & 0.470 & 0.489 \\
           & \AdvSimabbr & 0.570 & 0.350 & 0.650 & 0.900 & 0.600 & 0.820 & 0.520 & 0.550 & 0.620 \\
           & \CarlaGeneratorabbr & 0.610 & 0.630 & 0.322 & 0.900 & 0.767 & 0.756 & 0.667 & 0.711 & 0.670 \\
           & \AdvTrajabbr & 0.680 & 0.310 & 0.700 & 0.930 & 1.000 & 0.850 & 0.500 & 0.900 & \textbf{0.734} \\
        \midrule
           \multirow{4}{*}{S-\collisionrateabbr $\uparrow$} & \Learningtocollideabbr & 0.756 & 0.923 & 0.560 & 0.919 & 0.833 & 0.870 & 0.661 & 0.793 & 0.789 \\
           & \AdvSimabbr & 0.794 & 0.595 & 0.650 & 0.900 & 0.833 & 0.930 & 0.792 & 0.797 & 0.787 \\
           & \CarlaGeneratorabbr & 0.967 & 0.684 & 0.322 & 0.900 & 0.932 & 0.870 & 0.711 & 0.797 & 0.773 \\
           & \AdvTrajabbr & 0.847 & 0.485 & 0.697 & 0.930 & 1.000 & 0.966 & 0.562 & 1.000 & \textbf{0.811} \\
        \midrule
          \multirow{4}{*}{\overallscoreabbr $\downarrow$} & \Learningtocollideabbr & 0.765 & 0.825 & 0.613 & 0.451 & 0.755 & 0.632 & 0.630 & 0.646 & 0.665 \\
           & \AdvSimabbr & 0.654 & 0.718 & 0.577 & 0.465 & 0.659 & 0.544 & 0.599 & 0.606 & 0.603 \\
           & \CarlaGeneratorabbr & 0.629 & 0.577 & 0.738 & 0.464 & 0.569 & 0.571 & 0.520 & 0.522 & 0.574 \\
           & \AdvTrajabbr & 0.600 & 0.737 & 0.557 & 0.455 & 0.460 & 0.526 & 0.607 & 0.423 & \textbf{0.546} \\
        \midrule
          \multirow{4}{*}{S-\overallscoreabbr $\downarrow$} & \Learningtocollideabbr & 0.565 & 0.461 & 0.613 & 0.451 & 0.533 & 0.518 & 0.528 & 0.476 & 0.518 \\
           & \AdvSimabbr & 0.548 & 0.600 & 0.577 & 0.465 & 0.535 & 0.492 & 0.451 & 0.480 & 0.518 \\
           & \CarlaGeneratorabbr & 0.465 & 0.550 & 0.738 & 0.464 & 0.483 & 0.519 & 0.496 & 0.473 & 0.524 \\
           & \AdvTrajabbr & 0.523 & 0.654 & 0.558 & 0.455 & 0.460 & 0.471 & 0.574 & 0.372 & \textbf{0.508} \\
        \midrule
           \multirow{4}{*}{\selectionrateabbr $\uparrow$} & \Learningtocollideabbr & 0.410 & 0.130 & 1.000 & 0.990 & 0.420 & 0.690 & 0.590 & 0.580 & 0.601 \\
           & \AdvSimabbr & 0.680 & 0.420 & 1.000 & 1.000 & 0.720 & 0.860 & 0.530 & 0.640 & 0.731 \\
           & \CarlaGeneratorabbr & 0.600 & 0.760 & 1.000 & 1.000 & 0.822 & 0.856 & 0.922 & 0.878 & \textbf{0.855} \\
           & \AdvTrajabbr & 0.590 & 0.330 & 0.990 & 1.000 & 1.000 & 0.870 & 0.890 & 0.900 & 0.821 \\
        \bottomrule
    \end{tabular}
  }
}
\end{table}

\textbf{Analysis of generation algorithms and testing scenarios.} 
We analyze the properties of scenario generation algorithms based on a range of metrics, including the \textit{\collisionrate} (\collisionrateabbr), \textit{\overallscore} (\overallscoreabbr), and the overall \textit{\selectionrate} (\selectionrateabbr) for each scenario before and after selection.
As shown in \Cref{tab:generation_filtering}, first, the scenario selection process indeed helps to improve \collisionrateabbr of the testing scenarios to induce more safety-critical ones: with the highest improvement as $30\%$ for \Learningtocollideabbr.
Second, \AdvTrajabbr is the most effective algorithm to cause both high \collisionrateabbr and low \overallscoreabbr. In fact, $73.4\%$ of the generated scenarios by \AdvTrajabbr can cause collisions to the surrogate model, and it will increase to $81.1\%$ after scenario selection. The scenarios generated by \AdvTrajabbr achieve \overallscoreabbr as $0.546$, and it will further decrease to $0.508$ after scenario selection, indicating its testing effectiveness.
Third, regarding the overall \selectionrateabbr of different algorithms, scenarios generated by \CarlaGeneratorabbr achieve the highest \selectionrateabbr, which means \CarlaGeneratorabbr is the best algorithm in terms of transferability across different AD algorithms. Specifically, $85.5\%$ of scenarios generated by \CarlaGeneratorabbr can successfully cause collisions to other unseen AD agents.
Finally, among different scenarios, \textit{\Sfour} is the most difficult  with the highest \collisionrateabbr and  lowest \overallscoreabbr.

\subsection{Benchmark results}

We train our AD algorithms on random benign scenarios and evaluate them on \name. We present the training details in \Cref{appendix:agent_training}
and we provide important findings in the following.

\textbf{Performance of AD on benign and safety-critical scenarios.}
The benchmark results of AD algorithms based on 4D inputs are summarized in \Cref{tab:benchmark}. 
From the table, we observe a large performance gap in AD algorithms tested on benign and safety-critical scenarios in \name. 
For example, although TD3 achieves an \overallscore of $0.830$ on benign scenarios, it only achieves $0.518$ when testing on safety-critical scenarios.
In general, agents that perform well in benign scenarios usually fail given the safety-critical ones, indicating a trade-off between the performance under benign and safety-critical testing scenarios. 
For instance, PPO obtains the highest \overallscore on safety-critical scenarios, while its benign performance is worse than both SAC and TD3. 
On the other hand, although SAC achieves the highest \overallscore on benign testing scenarios, its performance under safety-critical ones is the worst.
More results on algorithms with other types of input observations can be found in \Cref{appendix:full_benchmark}.

 
\begin{table}[t]
\small
    \centering
    \caption{\small \textbf{The performance of AD algorithms on \name}. We report the average \textit{\overallscore} (\overallscoreabbr) on testing scenarios generated by all the $4$ scenario generation algorithms with driving route variations. \textit{Benign} indicates the performance of AD algorithms tested on normal driving scenarios. The last two columns show the \overallscoreabbr averaged over all benign and safety-critical traffic scenarios.}
    \label{tab:benchmark}
    \setlength{\tabcolsep}{3.75pt}
    \begin{tabular}{l|cccccccc|cc}
    \toprule
    \multirow{3}{*}{\textbf{Model}} & \multicolumn{8}{c|}{\textbf{Traffic Scenarios}} & \multirow{1}{*}{\textbf{Avg.}} & \multirow{1}{*}{\textbf{Avg.}} \\
    & \scriptsize{\makecell{Straight \\ Obstacle}} & \scriptsize{\makecell{Turning \\ Obstacle}} & \scriptsize{\makecell{Lane \\ Changing}}  & \scriptsize{\makecell{Vehicle \\ Passing}} & \scriptsize{\makecell{Red-light \\ Running}} & \scriptsize{\makecell{Unprotected \\ Left-turn}} & \scriptsize{\makecell{Right-\\ turn}} & \scriptsize{\makecell{Crossing \\ Negotiation}}
    & \scriptsize{Benign} & \scriptsize{\makecell{Safety-\\critical}} \\
    \midrule
    DDPG (4D) & 0.545 & 0.526 & 0.440 & 0.501 & 0.611 & 0.444 & 0.411 & 0.507 & 0.603 & 0.498 \\
    SAC (4D) & 0.533 & 0.474 & \textbf{0.577} & 0.471 & 0.482 & 0.501 & 0.503 & 0.432 & \textbf{0.833} & 0.497 \\
    TD3 (4D) & 0.479 & 0.596 & 0.477 & \textbf{0.592} & 0.532 & 0.525 & 0.459 & 0.482 & 0.830 & 0.518 \\
    PPO (4D) & \textbf{0.761} & \textbf{0.611} & 0.426 & 0.432 & \textbf{0.755} & \textbf{0.728} & \textbf{0.605} & \textbf{0.655} & 0.819 & \textbf{0.622} \\
    \bottomrule
    \end{tabular}
\end{table}

\textbf{Comprehensive diagnostic report of AD algorithms in all scenarios.}
In order to provide a comprehensive understanding of the performance of AD algorithms, we conducted a detailed diagnostic report for each tested algorithm from different perspectives. 
In particular, we consider three levels of evaluation metrics: Safety, Functionality, and Etiquette, as shown in \Cref{tab:report} for the 4D-based AD agents.
Comprehensive reports of all AD agents  are in \Cref{appendix:full_report}.
We observe that different AD algorithms outperform others under different metrics.
For instance, on the \textit{Safety level}, PPO achieves the lowest \collisionrateabbr and \outofroadabbr, which means it has a high level of safety and a low accident rate, while its performance on the Etiquette level is relatively low. 
On the \textit{Functionality level}, TD3 achieves the highest \followroute, demonstrating its ability to complete given tasks without deviating from the route. 
On the \textit{Etiquette level}, SAC and DDPG achieve the lowest \accelerationabbr and \yawvelocityabbr respectively, which measure the driving quality.
Based on the \overallscore (\overallscoreabbr), PPO is shown to be the best AD algorithm given the weighted average over all metrics.

We also notice a trade-off between functionality-level metrics and safety-level metrics. From \Cref{tab:report}, we can observe that an agent with strong functionality performance may not be safe regarding the safety level metrics. For instance, the SAC agent achieves the best \timespentabbr score, which means that it can finish the routes in the shortest time, but its collision rate (\collisionrateabbr) is also the highest among all the other agents. Similarly, the PPO agent that achieves the best route completion (\routecompletionabbr) score presents, however, the highest \runredlightabbr and \runstopsignabbr scores, which means that it may run red lights and stop signs most frequently. 
This observation suggests the inherent contradiction between some safety metrics and functionality metrics, which is also unveiled in some previous studies~\citep{ray2019benchmarking, liu2020constrained, liu2022robustness}.

\begin{table}[t]
\small
    \centering
    \caption{\small \textbf{Diagnostic report}. We test every AD algorithm on all selected testing scenarios and report the evaluation results on three different levels. \collisionrateabbr: \collisionrate, \runredlightabbr: \runredlight, \runstopsignabbr: \runstopsign, \outofroadabbr: \outofroad, \followrouteabbr: \followroute, \routecompletionabbr: \routecompletion, \timespentabbr: \timespent, \accelerationabbr: \acceleration, \yawvelocityabbr: \yawvelocity, \laneinvasionabbr: \laneinvasion, \overallscoreabbr: \overallscore, $\uparrow$/$\downarrow$: higher/lower the better.}
    \label{tab:report}
    \setlength{\tabcolsep}{3.75pt}
    \begin{tabular}{l|cccc|ccc|ccc|c}
    \toprule
    \multirow{2}{*}{\textbf{Model}} & \multicolumn{4}{c|}{\textbf{Safety Level}} & \multicolumn{3}{c|}{\textbf{Functionality Level}} & \multicolumn{3}{c|}{\textbf{Etiquette Level}} & \multirow{2}{*}{\textbf{\overallscoreabbr $\uparrow$}} \\
    & \collisionrateabbr $\downarrow$ & \runredlightabbr $\downarrow$ & \runstopsignabbr $\downarrow$ & \outofroadabbr $\downarrow$ & \followrouteabbr $\uparrow$ & \routecompletionabbr $\uparrow$ & \timespentabbr $\downarrow$ & \accelerationabbr $\downarrow$ & \yawvelocityabbr $\downarrow$ & \laneinvasionabbr $\downarrow$ & \\
    \midrule
    DDPG (4D) & 0.780 & \textbf{0.089} & \textbf{0.087} & 12.619 & 0.504 & 0.466 & 20.860 & 2.488 & \textbf{0.405} & 5.764 & 0.489 \\
    SAC (4D) & 0.829 & 0.216 & 0.146 & 3.115 & 0.882 & 0.648 & \textbf{16.827} & \textbf{1.830} & 0.704 & 2.580 & 0.499 \\
    TD3 (4D) & 0.783 & 0.231 & 0.141 & 2.535 & \textbf{0.903} & 0.670 & 17.644 & 2.680 & 1.493 & \textbf{2.545} & 0.516 \\
    PPO (4D) & \textbf{0.603} & 0.287 & 0.150 & \textbf{0.099} & 0.901 & \textbf{0.751} & 18.021 & 2.461 & 1.506 & 3.528 & \textbf{0.606} \\
    \bottomrule
    \end{tabular}
\end{table}



\subsection{Robustness Evaluation: Physical semantic attacks against AD algorithms}

\m{With our modularized design, \name is also able to identify  vulnerabilities of different components in AD systems by performing diverse \textit{adversarial attacks} in addition to the tests on safety-critical scenarios.
Here we provide evaluations against physical semantic attacks on perception components in AD such as LiDAR and multi-sensor systems, considering both the semantic segmentation and 3D object detection tasks.
More evaluations and visualizations are in~\Cref{appendix:toolkit}.}
     
\begin{table}[t]
\small
    \centering
    \caption{\small \m{\textbf{Robustness of point cloud segmentation}. 
    We test 4 point cloud segmentation models against three adversarial attacks  and report the IoU.
    }}
    \label{tab:lidar_attack}
    \setlength{\tabcolsep}{3.75pt}
      \begin{tabular}{c|c|c|c|c}
        \toprule
        Method            & PointNet++~\cite{qi2017pointnet++}        & SqueezeSeg~\cite{xu2020squeezesegv3}         & PolarSeg~\cite{zhang2020polarnet}       & Cylinder3D~\cite{zhou2020cylinder3d} \\
        \midrule
        Benign            &  0.81 $\pm$ 0.01      &  0.82 $\pm$ 0.01      & 0.95 $\pm$ 0.01     & 0.96 $\pm$ 0.01 \\
        \midrule
        Point Attack      &  0.80 $\pm$ 0.01      &  0.82 $\pm$ 0.02      & 0.94 $\pm$ 0.01     & 0.96 $\pm$ 0.01 \\
        Pose Attack       &  0.40 $\pm$ 0.08      &  0.47 $\pm$ 0.04      & 0.89 $\pm$ 0.01     & 0.88 $\pm$ 0.02 \\
        Scene Attack      &  0.52 $\pm$ 0.12      &  0.65 $\pm$ 0.04      & 0.85 $\pm$ 0.01     & 0.86 $\pm$ 0.01 \\
        \bottomrule
      \end{tabular}
\end{table}

\paragraph{\m{Point cloud segmentation}}
\m{To test the robustness of point cloud segmentation in AD, we implement 3 types of adversarial attacks: (1) \textit{Point Attack}: a point-wise attack method \cite{xiang2019generating} that adds small disturbance to the positions of 3D points; (2) \textit{Pose Attack}: a scene generation method that searches for adversarial poses of vehicles; (3) \textit{Scene Attack}: a semantically controllable generative method based on SAG~\cite{ding2021semantically}.
}
\m{
For Pose attack and Scene Attack, we first generate the locations and orientations of vehicles and spawn them in Carla. Then, we use the LiDAR sensor to collect point clouds 
needed by the segmentation algorithms. 
}
\m{We select 4 segmentation models (PointNet++~\cite{qi2017pointnet++}, PolarSeg~\cite{zhang2020polarnet}, SqueezeSegV3 \cite{xu2020squeezesegv3}, Cylinder3D~\cite{zhou2020cylinder3d}) as our victim models, all of which are pre-trained on Semantic Kitti dataset~\cite{behley2019iccv}. 
We present the evaluation results of IoU after attacks in~\Cref{tab:lidar_attack}. 
The results show that all 4 models can be attacked and have very different performances under attacks. 
This demonstrates the ability of \name on evaluating point cloud perception task in AD systems. 
}

\paragraph{\m{3D object detection}}
\m{To evaluate the robustness of recognizing and locating surrounding objects for AD systems, we perform different attacks on 3D object detection task.
We place the AD agent in a fixed scene and put different objects such as vehicles, pedestrians, and other traffic objects in front of the agent to test the detection accuracy. 
We follow TSS~\citep{li2021tss} to perform adversarial physical semantic transformations to both camera image and LiDAR point clouds. 
We incorporate $4$ kinds of semantic transformations to attack the perception component in AD systems. 
Specifically, we first consider changing different types of vehicles such as \textit{Tesla Model 3}, \textit{Audi TT}, and \textit{Nissan Patrol}. 
Second, we perturb the color of each vehicle. 
We  choose $5$ most common colors to test the robustness of AD algorithms and any RGB value can be applied to the car in \name.
Third, we change different properties for pedestrians, such as body shapes and skin colors. 
Finally, we perform rotation on every object to examine the reliability of AD systems.
We present our results in~\Cref{tab:adv_attack}. 
We train  different 3D object detection models on normal driving scenarios and test the models on adversarial data generated by \name. 
The SECOND~\citep{yan2018second} takes LiDAR point clouds as input while CLOCs~\citep{pang2020clocs} is a multi-modal model that takes both LiDAR point clouds and camera images as inputs. 
From \Cref{tab:adv_attack}, we find that the SECOND model performs better on benign data. 
However, on adversarial data, CLOCs achieves higher average precision and the performance drop of the CLOCs model from benign to adversarial is much smaller than that of the SECOND model. 
One reason could be that data from both modalities complement each other, helping the model to make better decisions, which indicates potential designing strategies for AD algorithms: use multi-sensor fusion models to incorporate and process multi-modal data, leading to higher robustness. 
This demonstrates the ability of \name to evaluate the robustness of object detection task. 
}

\begin{table}[t]
\small
    \centering
    \caption{\small \m{\textbf{Robustness of 3D object detection.} We report the average precision (AP) of car class for models taking 3D LiDAR point clouds and multi-modal data as inputs respectively.}}
    \label{tab:adv_attack}
    \setlength{\tabcolsep}{3.75pt}
    \begin{tabular}{l|l|l|ccc|ccc}
    \toprule
    \multirow{2}{*}{\textbf{Data Source}} & \multirow{2}{*}{\textbf{Model}} & \multirow{2}{*}{\textbf{Input Data}} & \multicolumn{3}{c|}{\textbf{3D AP (\%)}} & \multicolumn{3}{c}{\textbf{Bird’s Eye View AP (\%)}} \\
    & & & easy & moderate & hard & easy & moderate & hard \\
    \midrule
    \multirow{2}{*}{Benign} & SECOND & LiDAR & \textbf{87.31} & \textbf{86.81} & \textbf{86.81} & \textbf{88.95} & \textbf{88.92} & \textbf{88.92} \\
    & CLOCs & LiDAR+Img & 76.90 & 76.50 & 76.50 & 81.73 & 82.01 & 82.01 \\
    \midrule
    \multirow{2}{*}{Adversarial} & SECOND & LiDAR & 60.74 & 59.81 & 59.81 & 67.87 & 63.64 & 63.64 \\
    & CLOCs & LiDAR+Img & \textbf{61.98} & \textbf{61.98} & \textbf{61.98} & \textbf{76.64} & \textbf{76.64} & \textbf{76.64} \\
    \bottomrule
    \end{tabular}
\end{table}

\section{Conclusion}
\label{sec:conclusion}
In this paper, we introduce \name, the first unified platform to automatically evaluate and analyze the performance of AD algorithms in multiple aspects using various safety-critical driving scenarios generated by different generation algorithms. 
We incorporate $8$ safety-critical scenarios and $10$ evaluation metrics from $3$ different levels to provide a detailed diagnostic report for each AD agent.
AD algorithms tested on \name have a large performance drop compared to evaluations on benign scenarios, suggesting the deficiencies of each algorithm and the effectiveness of our testing platform.
We hope our platform and findings will serve as a reliable and comprehensive benchmark to help researchers and practitioners to identify weaknesses in existing AD systems and further develop safe AD algorithms as well as more effective testing scenario generation algorithms.

\paragraph{\m{Limitations}}
\m{Although simulation is a useful and necessary tool for evaluating AD systems given its efficiency and controllability~\citep{amini2022vista, thorn2018framework}, the simulation in \name cannot exactly reflect real-world conditions. 
On-track testing is necessary before deploying AD algorithms in the real world. 
Besides, we only evaluate RL-based AD algorithms in the current version of SafeBench, and testing more diverse AD algorithms, including commercial systems such as Baidu Apollo~\citep{apollosimulation} and Openpilot~\citep{openpilot} would be interesting future work.}

\begin{ack}
\vspace{-2mm}
This work is partially supported by the NSF grant No.1910100,
NSF CNS No.2046726, C3 AI, and the Alfred P. Sloan Foundation.
\end{ack}

\bibliographystyle{unsrt}
\bibliography{main}

\clearpage
\section*{Checklist}


\begin{enumerate}

\item For all authors...
\begin{enumerate}
  \item Do the main claims made in the abstract and introduction accurately reflect the paper's contributions and scope?
    \answerYes{See \Cref{sec:intro}.}
  \item Did you describe the limitations of your work?
    \answerYes{See \Cref{sec:conclusion}}
  \item Did you discuss any potential negative societal impacts of your work?
    \answerYes{See \Cref{appendix:impact}}
  \item Have you read the ethics review guidelines and ensured that your paper conforms to them?
    \answerYes{}
\end{enumerate}

\item If you are including theoretical results...
\begin{enumerate}
  \item Did you state the full set of assumptions of all theoretical results?
    \answerNA{}
	\item Did you include complete proofs of all theoretical results?
    \answerNA{}
\end{enumerate}

\item If you ran experiments (e.g. for benchmarks)...
\begin{enumerate}
  \item Did you include the code, data, and instructions needed to reproduce the main experimental results (either in the supplemental material or as a URL)?
    \answerYes{See on the website: \url{https://safebench.github.io}.}
  \item Did you specify all the training details (e.g., data splits, hyperparameters, how they were chosen)?
    \answerYes{See \Cref{appendix:agent_training}}
	\item Did you report error bars (e.g., with respect to the random seed after running experiments multiple times)?
    \answerNA{}
	\item Did you include the total amount of compute and the type of resources used (e.g., type of GPUs, internal cluster, or cloud provider)?
    \answerYes{See \Cref{appendix:agent_training}}
\end{enumerate}

\item If you are using existing assets (e.g., code, data, models) or curating/releasing new assets...
\begin{enumerate}
  \item If your work uses existing assets, did you cite the creators?
    \answerYes{See \Cref{sec:scenario_generation}}
  \item Did you mention the license of the assets?
    \answerYes{See on the website: \url{https://safebench.github.io}.}
  \item Did you include any new assets either in the supplemental material or as a URL?
    \answerYes{See on the website: \url{https://safebench.github.io}.}
  \item Did you discuss whether and how consent was obtained from people whose data you're using/curating?
    \answerNA{}
  \item Did you discuss whether the data you are using/curating contains personally identifiable information or offensive content?
    \answerYes{See \Cref{appendix:full_generation}}
\end{enumerate}

\item If you used crowdsourcing or conducted research with human subjects...
\begin{enumerate}
  \item Did you include the full text of instructions given to participants and screenshots, if applicable?
    \answerNA{}
  \item Did you describe any potential participant risks, with links to Institutional Review Board (IRB) approvals, if applicable?
    \answerNA{}
  \item Did you include the estimated hourly wage paid to participants and the total amount spent on participant compensation?
    \answerNA{}
\end{enumerate}

\end{enumerate}


\clearpage

\appendix


\section{Appendix}
\subsection{\name statistics}
\label{appendix:scenario_statistics}

We present the statistics of testing scenarios generated by each generation algorithm in \Cref{tab:statistics}. 
For each algorithm, we report the statistics both before and after scenario selection, where we only keep scenarios that have high transferability across AD algorithms. 
By applying the $4$ generation algorithms, we obtain \rawscenarionumber testing scenarios in total, from which we select \finalscenarionumber testing scenarios for AD evaluation.

\m{\subsection{\name design details}}

\m{Our evaluation platform runs in the Docker container and is built upon the Carla simulator~\citep{dosovitskiy2017carla}. 
We design 4 components (nodes) that are highly flexible for users to customize: ego vehicle, agent node, scenario node, and evaluation node. These components communicate with each other through ROS. We detail the platform design as follows.}

\paragraph{\m{Docker image}}

\m{We provide a docker image containing \name, which makes the platform more portable. The docker image is built based on Ubuntu 20.04. Inside the docker image, we have pre-installed Carla 0.9.11 and ROS Noetic for simulation and communication in \name respectively.}

\paragraph{\m{ROS services}}

\m{The communication between different nodes is implemented using ROS services. For example, when the AD algorithm in the agent node is ready, it will request the waypoint information specified by the scenario node. The scenario node will send out the waypoint information in the current scenario once it receives a request from the agent node.}

\paragraph{\m{CARLA}}

\m{We use Carla 0.9.11 as our traffic simulator. The scenario runner~\citep{scenariorunner} is incorporated in the scenario node to easily define and execute different scenarios. In the agent node, we develop our RL agent based on Gym Carla~\citep{gymcarla} environment which supports an OpenAI gym style interaction between the agent and Carla simulator. }

\subsection{Definition of scenarios and examples of route variants}
\label{appendix:scenario_route}

\begin{figure}[t]
     \centering
     \begin{subfigure}[t]{0.48\textwidth}
         \centering
         \includegraphics[width=\textwidth]{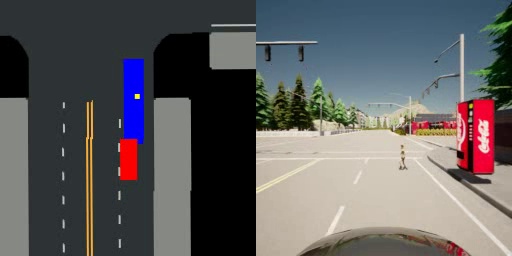}
         \caption{\Sone}
         \label{subfig:scenario1}
     \end{subfigure}
     \hfill
     \begin{subfigure}[t]{0.48\textwidth}
         \centering
         \includegraphics[width=\textwidth]{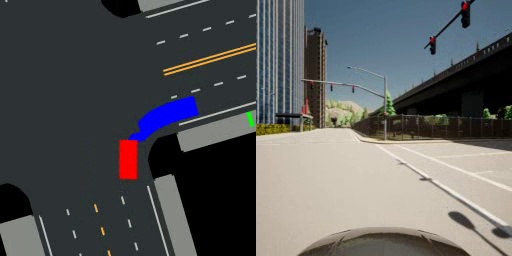}
         \caption{\Stwo}
         \label{subfig:scenario2}
     \end{subfigure}
     \hfill
     \begin{subfigure}[t]{0.48\textwidth}
         \centering
         \includegraphics[width=\textwidth]{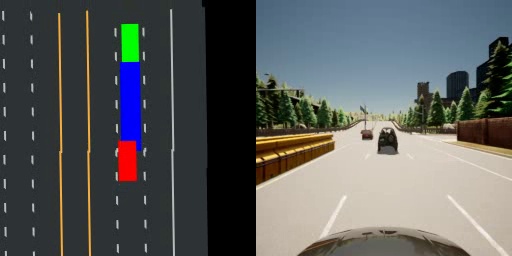}
         \caption{\Sthree}
         \label{subfig:scenario3}
     \end{subfigure}
     \hfill
     \begin{subfigure}[t]{0.48\textwidth}
         \centering
         \includegraphics[width=\textwidth]{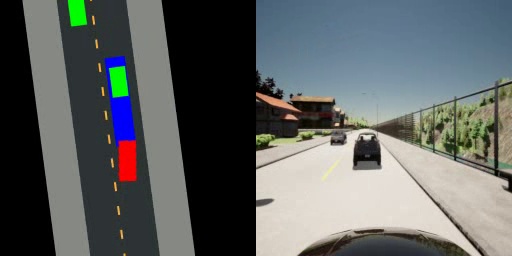}
         \caption{\Sfour}
         \label{subfig:scenario4} 
     \end{subfigure}
     \hfill
     \begin{subfigure}[t]{0.48\textwidth}
         \centering
         \includegraphics[width=\textwidth]{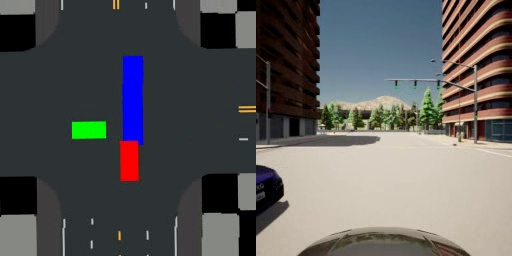}
         \caption{\Sfive}
         \label{subfig:scenario5}
     \end{subfigure}
     \hfill
     \begin{subfigure}[t]{0.48\textwidth}
         \centering
         \includegraphics[width=\textwidth]{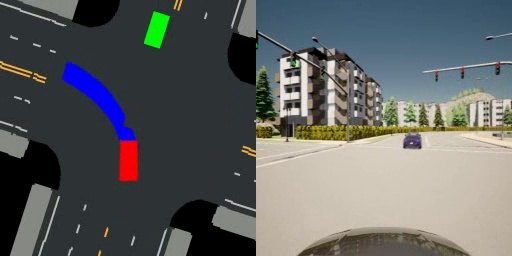}
         \caption{\Ssix}
         \label{subfig:scenario6}
     \end{subfigure}
     \hfill
     \begin{subfigure}[t]{0.48\textwidth}
         \centering
         \includegraphics[width=\textwidth]{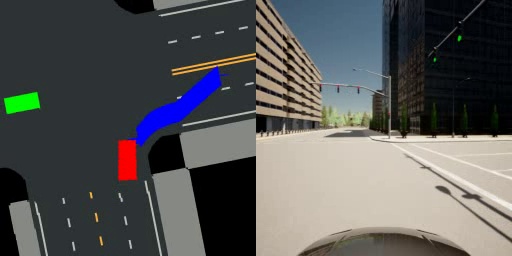}
         \caption{\Sseven}
         \label{subfig:scenario7}
     \end{subfigure}
     \hfill
     \begin{subfigure}[t]{0.48\textwidth}
         \centering
         \includegraphics[width=\textwidth]{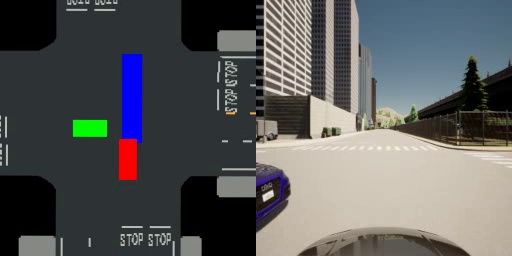}
         \caption{\Seight}
         \label{subfig:scenario8}
     \end{subfigure}
        \caption{Pre-crash scenarios.}
        \label{fig:scenarios}
\end{figure}

We first give detailed definitions of the $8$ traffic scenarios considered in \name in Table~\ref{tab:sceario_description} and screenshots of them in Figure~\ref{fig:scenarios}.
We also develop benign scenarios based on these safety-critical scenarios. In benign situations, everything is the same except that the other vehicles are auto-piloted. As a result, we have $8$ kinds of benign scenarios, and we can compare the benign performances with safety-critical ones.

\begin{figure}[t]
     \centering
     \begin{subfigure}[t]{0.32\textwidth}
         \centering
         \includegraphics[width=\textwidth]{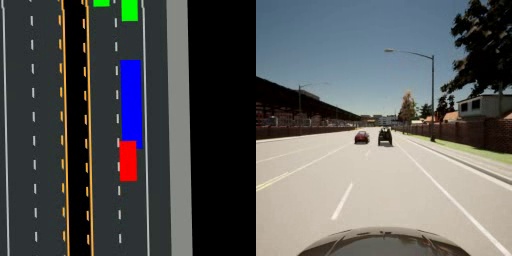}
         \caption{Two-lane highway}
         \label{subfig:scenario3_v1}
     \end{subfigure}
     \hfill
     \begin{subfigure}[t]{0.32\textwidth}
         \centering
         \includegraphics[width=\textwidth]{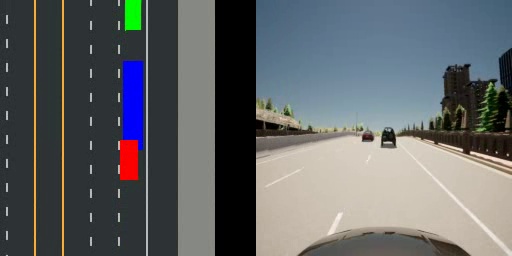}
         \caption{Three-lane bridge}
         \label{subfig:scenario3_v2}
     \end{subfigure}
     \hfill
     \begin{subfigure}[t]{0.32\textwidth}
         \centering
         \includegraphics[width=\textwidth]{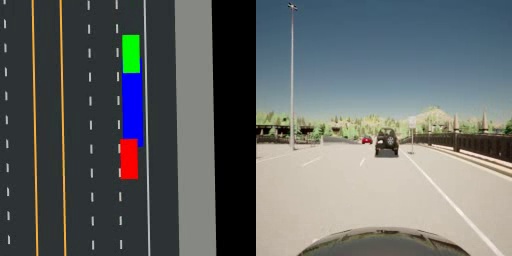}
         \caption{Three-lane bridge with a speed limit sign}
         \label{subfig:scenario3_v3}
     \end{subfigure}
        \caption{Example route variants of scenario 3.}
        \label{fig:variants_s3}
\end{figure}

\begin{figure}[!htb]
     \centering
     \begin{subfigure}[t]{0.32\textwidth}
         \centering
         \includegraphics[width=\textwidth]{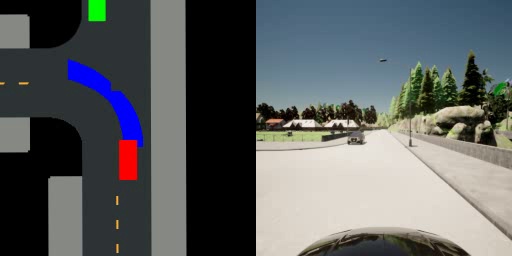}
         \caption{Single-lane T-junction without surrounding buildings}
         \label{subfig:scenario6_v1}
     \end{subfigure}
     \hfill
     \begin{subfigure}[t]{0.32\textwidth}
         \centering
         \includegraphics[width=\textwidth]{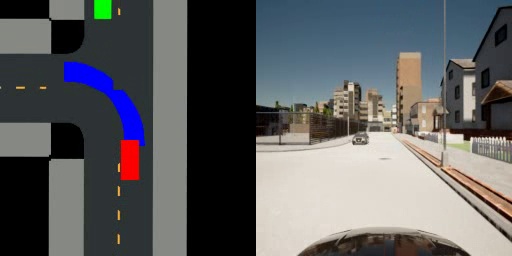}
         \caption{Single-lane T-junction with surrounding buildings}
         \label{subfig:scenario6_v2}
     \end{subfigure}
     \hfill
     \begin{subfigure}[t]{0.32\textwidth}
         \centering
         \includegraphics[width=\textwidth]{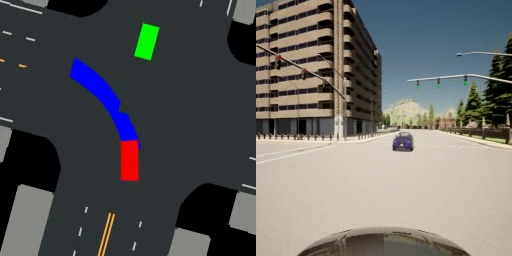}
         \caption{Two-lane intersection}
         \label{subfig:scenario6_v3}
     \end{subfigure}
        \caption{Example route variants of scenario 6.}
        \label{fig:variants_s6}
\end{figure}

\begin{figure}[!htb]
     \centering
     \begin{subfigure}[t]{0.32\textwidth}
         \centering
         \includegraphics[width=\textwidth]{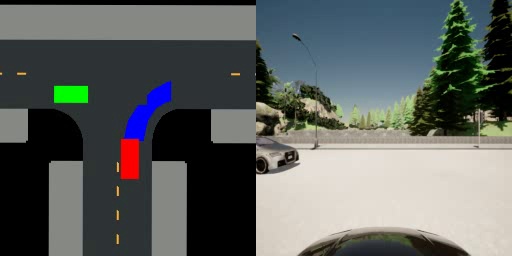}
         \caption{Single-lane T-junction}
         \label{subfig:scenario7_v1}
     \end{subfigure}
     \hfill
     \begin{subfigure}[t]{0.32\textwidth}
         \centering
         \includegraphics[width=\textwidth]{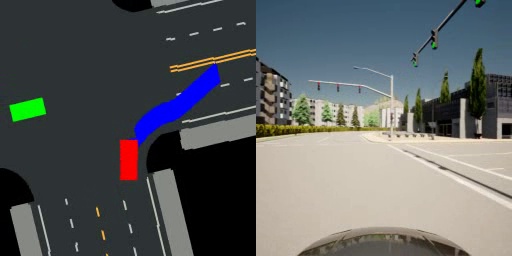}
         \caption{Two-lane intersection}
         \label{subfig:scenario7_v2}
     \end{subfigure}
     \hfill
     \begin{subfigure}[t]{0.32\textwidth}
         \centering
         \includegraphics[width=\textwidth]{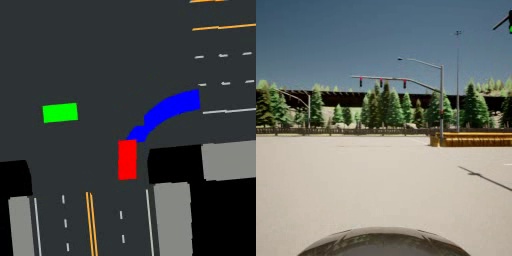}
         \caption{Two-lane T-junction}
         \label{subfig:scenario7_v3}
     \end{subfigure}
        \caption{Example route variants of scenario 7.}
        \label{fig:variants_s7}
\end{figure}

We show more examples of route variants incorporated in our evaluation platform in \Cref{fig:variants_s3,,fig:variants_s6,,fig:variants_s7}.

\begin{table}[t]
\small
\centering
    \caption{Scenario Description}
    \label{tab:sceario_description}
{
{
\setlength{\tabcolsep}{3.75pt}
    \begin{tabular}{l|p{10cm}}
    \toprule
        \textbf{Scenario Name} & Description \\
        \midrule
        \Sone & 
        The ego vehicle encounters an unexpected cyclist or pedestrian on the road and must perform an emergency brake or an avoidance maneuver. As shown in \Cref{subfig:scenario1}, the vision of the ego vehicle is usually blocked by an obstacle, which is safety-critical since the reaction time left for the ego vehicle is very short. \\
        \midrule
        \Stwo &
        As shown in \Cref{subfig:scenario2}, while turning at an intersection, the ego vehicle finds an unexpected cyclist or pedestrian on the road and must perform an emergency brake or an avoidance maneuver. \\
        \midrule
        \Sthree &
        In this scenario, the ego vehicle should perform a lane changing to evade a leading vehicle, which is moving too slowly. In addition, there is another leading vehicle in the adjacent lane, which is traveling at a normal speed. The ego vehicle needs to avoid hitting both cars when overtaking. See \Cref{subfig:scenario3} for more details. \\
        \midrule
        \Sfour & 
        The ego vehicle must go around a blocking object using the opposite lane, dealing with oncoming traffic. The ego vehicle should avoid colliding with both cars and also avoid driving outside the lane. We provide an example in \Cref{subfig:scenario4}. \\
        \midrule
        \Sfive & 
        When the ego vehicle is going straight at an intersection, a crossing vehicle runs a red light. The ego vehicle is forced to take actions to avoid potential collisions as shown in \Cref{subfig:scenario5}. \\
        \midrule
        \Ssix &
        As shown in \Cref{subfig:scenario6}, the ego vehicle is performing an unprotected left turn at an intersection while there is a vehicle going straight in the opposite lane. \\
        \midrule
        \Sseven & 
        In this scenario, the ego vehicle is performing a right turn at an intersection, with a crossing vehicle in front.  Collision avoidance actions must be taken to keep safe. We present an example in \Cref{subfig:scenario7}. \\
        \midrule
        \Seight &
        In this scenario, the ego vehicle meets another crossing vehicle when passing an intersection with no traffic lights. As shown in \Cref{subfig:scenario8}, the ego vehicle should negotiate with the other vehicle to cross the unsignalized intersection in an orderly and safe manner. \\
        \bottomrule
    \end{tabular}
  }
}
\end{table}

\begin{table}[t]
\small
\centering
    \caption{Statistics of \name testing scenarios.}
    \label{tab:statistics}
{
{
\setlength{\tabcolsep}{3.75pt}
    \begin{tabular}{ll|cccccccc|c}
    \toprule
        \multirow{3}{*}{\textbf{Algo.}} & \multirow{3}{*}{\textbf{\makecell{Scenario \\ Selection}}} & \multicolumn{8}{c|}{\textbf{Traffic Scenarios}} & \multirow{3}{*}{\textbf{Total}} \\
        & & \scriptsize{\makecell{Straight \\ Obstacle}} & \scriptsize{\makecell{Left-turn \\ Obstacle}} & \scriptsize{\makecell{Lane \\ Changing}}  & \scriptsize{\makecell{Vehicle \\ Passing}} & \scriptsize{\makecell{Red-light \\ Running}} & \scriptsize{\makecell{Unprotected \\ Left-turn}} & \scriptsize{\makecell{Right-turn}} & \scriptsize{\makecell{Crossing \\ Negotiation}} & \\
        \midrule
          \multirow{2}{*}{\Learningtocollideabbr} & \footnotesize{Before} & 100 & 100 & 100 & 100 & 100 & 100 & 100 & 100 & 800 \\
           & \footnotesize{After} & 41 & 13 & 100 & 99 & 42 & 69 & 59 & 58 & 481 \\
        \midrule
          \multirow{2}{*}{\AdvSimabbr} & \footnotesize{Before} & 100 & 100 & 100 & 100 & 100 & 100 & 100 & 100 & 800 \\
           & \footnotesize{After} & 68 & 42 & 100 & 100 & 72 & 86 & 53 & 64 & 585 \\
        \midrule
          \multirow{2}{*}{\CarlaGeneratorabbr} & \footnotesize{Before} & 100 & 100 & 90 & 90 & 90 & 90 & 90 & 90 & 740 \\
           & \footnotesize{After} & 60 & 76 & 90 & 90 & 74 & 77 & 83 & 79 & 629 \\
        \midrule
          \multirow{2}{*}{\AdvTrajabbr} & \footnotesize{Before} & 100 & 100 & 100 & 100 & 100 & 100 & 100 & 100 & 800 \\
           & \footnotesize{After} & 59 & 33 & 99 & 100 & 100 & 87 & 89 & 90 & 657 \\
        \midrule
          \multirow{2}{*}{Total} & \footnotesize{Before} & 400 & 400 & 390 & 390 & 390 & 390 & 390 & 390 & 3140 \\
           & \footnotesize{After} & 228 & 164 & 389 & 389 & 288 & 319 & 284 & 291 & 2352 \\
        \bottomrule
    \end{tabular}
  }
}
\end{table}

\subsection{Evaluation metrics}
\label{appendix:evaluation_metrics}

\begin{table}[t]
\small
\centering
    \caption{Constants and weights used in \name evaluation metrics.}
    \label{tab:constants}
{
{
\setlength{\tabcolsep}{3.75pt}
    \begin{tabular}{l|cccc|ccc|ccc}
    \toprule
        \multirow{2}{*}{\textbf{Symbol}} & \multicolumn{4}{c|}{\textbf{Safety Level}} & \multicolumn{3}{c|}{\textbf{Functionality Level}} & \multicolumn{3}{c}{\textbf{Etiquette Level}} \\
    & \collisionrateabbr & \runredlightabbr & \runstopsignabbr & \outofroadabbr & \followrouteabbr & \routecompletionabbr & \timespentabbr & \accelerationabbr & \yawvelocityabbr & \laneinvasionabbr \\
        \midrule
          $m^i_{max}$ & 1 & 1 & 1 & 50 & 1 & 1 & 60 & 8 & 3 & 20 \\
          $w^i$ & 0.495 & 0.099 & 0.099 & 0.099 & 0.050 & 0.050 & 0.050 & 0.020 & 0.020 & 0.020 \\
        \bottomrule
    \end{tabular}
  }
}
\end{table}

We follow the equations introduced in \Cref{sec:evaluation_metric} to calculate evaluation metrics. 
Specifically, for \textit{\followroute}, we first set $x_{max}$ to $5$ and then calculate the expectation. 
For other metrics, we directly calculate the expectation of each variable over the scenario distribution $\mathcal{P}$.
When calculating the \textit{\overallscore}, we follow the maximum allowed value $m^i_{max}$ and weights $w^i$ for each metric $m^i$ given in \Cref{tab:constants}. 
The weight for each metric depends on the evaluation level. 
Metrics in \textit{Safety Level} are assigned the highest weights since they focus on serious violations of traffic rules.
Among the $4$ safety level metrics, the weight of \collisionrateabbr is $5$ times larger than others' weights. 
The weights of metrics in \textit{Functionality Level} are one-half of the weights in \textit{Safety Level}, while the weights in \textit{Etiquette Level} are only one-fifth of them. 
Such a weight setup first emphasizes safety and then encourages the ego vehicle to complete the given tasks in a comfortable way.

\subsection{Implementation details of AD algorithms}
\label{appendix:agent_implementation}


\paragraph{Reward function}
During training, all RL algorithms share the same reward function.
The reward is a weighted sum of $7$ items.
We set the weight of longitudinal speed to $1$, the weight of lateral acceleration to $0.2$, and the weight of steering to $5$. 
If the ego vehicle encounters a collision or drives out of lane, we give a reward of $-1$ as a penalty.
If the speed of the ego vehicle is larger than a threshold, we give a reward of $-10$ as a penalty. 
The speed threshold is set to $9$.
We also add a constant reward of $0.1$.

\paragraph{Action space}
Similarly, the action space of every RL model is the same, which includes acceleration and a steering value.
For acceleration, the maximum and minimum allowed values are $3$ and $-3$, respectively. 
We limit the absolute value of steering to no greater than $0.3$. 
After having the acceleration and steering, we need to convert these values into Carla's vehicle control format, where we need to calculate the throttle and brake of the ego vehicle.
The throttle and brake are calculated using the following equations:
\begin{equation}
    throttle = 
    \begin{cases}
    acc / 3, & acc > 0 \\
    0, & otherwise \\
    \end{cases},
    brake = 
    \begin{cases}
    0, & acc > 0 \\
    - acc / 8, & otherwise \\
    \end{cases}
\end{equation}
where $acc$ denotes the acceleration given by RL models. Both throttle and brake will be clipped to the interval $[0, 1]$.

\paragraph{Model Architecture}
The model we used for deep RL methods is a simple multi-layer perceptron. 
The size of the hidden layer is [256, 256]. 
When adding bird-eye view images or camera images into input information, we use a separate image encoder to extract image features.
The encoder is end-to-end trained with the actor network in RL models.
We provide more details about the architecture of the image encoder in \Cref{tab:encoder}.

\begin{table}[t]
\small
\centering
    \caption{Model architecture of image encoder.}
    \label{tab:encoder}
{
{
\setlength{\tabcolsep}{3.75pt}
    \begin{tabular}{lccccc}
    \toprule
        \textbf{Layer} & \textbf{Input Channels} & \textbf{Output Channels} & \textbf{Kernel Size} & \textbf{Stride} & \textbf{Padding}\\
        \midrule
          Convolution Layer 1 & 3 & 32 & 3 & 2 & 1 \\
          Convolution Layer 2 & 32 & 64 & 3 & 2 & 1 \\
          Max Pooling Layer 1 & 64 & 64 & 3 & 3 & 0 \\
          Convolution Layer 3 & 64 & 128 & 3 & 2 & 1 \\
          Convolution Layer 4 & 128 & 256 & 3 & 2 & 1 \\
          Max Pooling Layer & 256 & 256 & 3 & 2 & 0 \\
          Fully Connect Layer 1 & 1024 & 512 & - & - & - \\
          Fully Connect Layer 2 & 512 & 256 & - & - & - \\
          Fully Connect Layer 3 & 256 & 128 & - & - & - \\
        \bottomrule
    \end{tabular}
  }
}
\end{table}

\paragraph{DDPG hyperparameters}
The policy learning rate is 0.0003, and the Q-value learning rate is 0.001. The standard deviation for Gaussian exploration noise added to the policy at training time is 0.1. The discount factor is 0.99. The number of models in the Q-ensemble critic is 1.

\paragraph{SAC hyperparameters}
The policy learning rate and Q-value learning rate are set to be 0.001. The entropy regularization coefficient, which is equivalent to the inverse of the reward scale in the original SAC paper, is 0.1. The discount factor equals 0.99, and the number of models in the Q-ensemble critic is 2.

\paragraph{TD3 hyperparameters}
The policy learning rate and Q-value learning rate are set to 0.001. The standard deviation for Gaussian exploration noise added to the policy at training time is 0.1. The standard deviation for smoothing noise added to noise is 0.2 The limit for the absolute value of smoothing noise is 0.5. Policy update delay is 2. The discount factor is 0.99. The number of models in the Q-ensemble critic is 2.

\paragraph{PPO hyperparameters}
The policy learning rate is 0.0003, and the Q-value learning rate is 0.001. The clip ratio of the policy object is 0.2. The target KL divergence is 0.01. We set both actor and critic training iterations to 80. The discount factor is 0.99, and the number of interaction steps is 1000.

\subsection{Training details of AD algorithms}
\label{appendix:agent_training}

All of the 4 deep RL algorithms are trained in Carla town03. 
Because town03 is the most complex town, with a 5-lane junction, a roundabout, unevenness, a tunnel, and more, according to Carla's official document.
The number of warm-up steps for off-policy methods is 600. 
The interpolation factor in polyak averaging for the target network is 0.995. 
The number of training epochs is different for different algorithms and different input states.
For example, SAC with 4D+Cam input is trained for $324$ epochs while DDPG with 4D input state is trained for $370$ epochs.
We train our RL models on NVIDIA GeForce RTX 3090 GPUs, and the training usually takes one day.
For each trained model, we achieve a stable reward value of around 1500 for one episode.

During scenario generation, we also train a SAC model with 4D input state space as a surrogate model. The training process is the same as other models except that we use a different random seed to produce a different training result.

\subsection{Detailed scenario generation results}
\label{appendix:full_generation}

\begin{table}[t]
\small
\centering
    \caption{Full statistics of scenario generation and selection.}
    \label{tab:generation_filtering_full}
{
{
\setlength{\tabcolsep}{3.75pt}
    \begin{tabular}{c|c|cccccccc|c}
    \toprule
        \multirow{3}{*}{\textbf{Metric}} & \multirow{3}{*}{\textbf{Algo.}} & \multicolumn{8}{c|}{\textbf{Traffic Scenarios}} & \multirow{3}{*}{\textbf{Avg.}} \\
        & & \scriptsize{\makecell{Straight \\ Obstacle}} & \scriptsize{\makecell{Turning \\ Obstacle}} & \scriptsize{\makecell{Lane \\ Changing}}  & \scriptsize{\makecell{Vehicle \\ Passing}} & \scriptsize{\makecell{Red-light \\ Running}} & \scriptsize{\makecell{Unprotected \\ Left-turn}} & \scriptsize{\makecell{Right-\\ turn}} & \scriptsize{\makecell{Crossing \\ Negotiation}} & \\
        \midrule
          \multirow{4}{*}{\collisionrateabbr $\uparrow$} & \Learningtocollideabbr & 0.320 & 0.140 & 0.560 & 0.920 & 0.410 & 0.630 & 0.458 & 0.470 & 0.489 \\
           & \AdvSimabbr & 0.570 & 0.350 & 0.650 & 0.900 & 0.600 & 0.820 & 0.520 & 0.550 & 0.620 \\
           & \CarlaGeneratorabbr & 0.610 & 0.630 & 0.322 & 0.900 & 0.767 & 0.756 & 0.667 & 0.711 & 0.670 \\
           & \AdvTrajabbr & 0.680 & 0.310 & 0.700 & 0.930 & 1.000 & 0.850 & 0.500 & 0.900 & \textbf{0.734} \\
        \midrule
           \multirow{4}{*}{S-\collisionrateabbr $\uparrow$} & \Learningtocollideabbr & 0.756 & 0.923 & 0.560 & 0.919 & 0.833 & 0.870 & 0.661 & 0.793 & 0.789 \\
           & \AdvSimabbr & 0.794 & 0.595 & 0.650 & 0.900 & 0.833 & 0.930 & 0.792 & 0.797 & 0.787 \\
           & \CarlaGeneratorabbr & 0.967 & 0.684 & 0.322 & 0.900 & 0.932 & 0.870 & 0.711 & 0.797 & 0.773 \\
           & \AdvTrajabbr & 0.847 & 0.485 & 0.697 & 0.930 & 1.000 & 0.966 & 0.562 & 1.000 & \textbf{0.811} \\
        \midrule
          \multirow{4}{*}{\routecompletionabbr $\downarrow$} & \Learningtocollideabbr & 0.842 & 0.934 & 0.704 & 0.680 & 0.805 & 0.744 & 0.843 & 0.780 & 0.792 \\
           & \AdvSimabbr & 0.713 & 0.928 & 0.649 & 0.673 & 0.740 & 0.646 & 0.827 & 0.762 & 0.742 \\
           & \CarlaGeneratorabbr & 0.693 & 0.874 & 0.886 & 0.674 & 0.656 & 0.666 & 0.760 & 0.680 & 0.736 \\
           & \AdvTrajabbr & 0.681 & 0.938 & 0.595 & 0.652 & 0.535 & 0.644 & 0.817 & 0.583 & \textbf{0.681} \\
        \midrule
           \multirow{4}{*}{S-\routecompletionabbr $\downarrow$} & \Learningtocollideabbr & 0.631 & 0.559 & 0.704 & 0.679 & 0.601 & 0.647 & 0.771 & 0.631 & 0.653 \\
           & \AdvSimabbr & 0.600 & 0.884 & 0.649 & 0.673 & 0.639 & 0.595 & 0.725 & 0.655 & 0.678 \\
           & \CarlaGeneratorabbr & 0.521 & 0.866 & 0.886 & 0.674 & 0.582 & 0.614 & 0.740 & 0.640 & 0.690 \\
           & \AdvTrajabbr & 0.576 & 0.905 & 0.596 & 0.652 & 0.535 & 0.594 & 0.794 & 0.536 & \textbf{0.649} \\
        \midrule
          \multirow{4}{*}{\overallscoreabbr $\downarrow$} & \Learningtocollideabbr & 0.765 & 0.825 & 0.613 & 0.451 & 0.755 & 0.632 & 0.630 & 0.646 & 0.665 \\
           & \AdvSimabbr & 0.654 & 0.718 & 0.577 & 0.465 & 0.659 & 0.544 & 0.599 & 0.606 & 0.603 \\
           & \CarlaGeneratorabbr & 0.629 & 0.577 & 0.738 & 0.464 & 0.569 & 0.571 & 0.520 & 0.522 & 0.574 \\
           & \AdvTrajabbr & 0.600 & 0.737 & 0.557 & 0.455 & 0.460 & 0.526 & 0.607 & 0.423 & \textbf{0.546} \\
        \midrule
          \multirow{4}{*}{S-\overallscoreabbr $\downarrow$} & \Learningtocollideabbr & 0.565 & 0.461 & 0.613 & 0.451 & 0.533 & 0.518 & 0.528 & 0.476 & 0.518 \\
           & \AdvSimabbr & 0.548 & 0.600 & 0.577 & 0.465 & 0.535 & 0.492 & 0.451 & 0.480 & 0.518 \\
           & \CarlaGeneratorabbr & 0.465 & 0.550 & 0.738 & 0.464 & 0.483 & 0.519 & 0.496 & 0.473 & 0.524 \\
           & \AdvTrajabbr & 0.523 & 0.654 & 0.558 & 0.455 & 0.460 & 0.471 & 0.574 & 0.372 & \textbf{0.508} \\
        \midrule
           \multirow{4}{*}{\selectionrateabbr $\uparrow$} & \Learningtocollideabbr & 0.410 & 0.130 & 1.000 & 0.990 & 0.420 & 0.690 & 0.590 & 0.580 & 0.601 \\
           & \AdvSimabbr & 0.680 & 0.420 & 1.000 & 1.000 & 0.720 & 0.860 & 0.530 & 0.640 & 0.731 \\
           & \CarlaGeneratorabbr & 0.600 & 0.760 & 1.000 & 1.000 & 0.822 & 0.856 & 0.922 & 0.878 & \textbf{0.855} \\
           & \AdvTrajabbr & 0.590 & 0.330 & 0.990 & 1.000 & 1.000 & 0.870 & 0.890 & 0.900 & 0.821 \\
        \bottomrule
    \end{tabular}
  }
}
\end{table}

We show the full scenario generation and selection statistics in \Cref{tab:generation_filtering_full}. 
We note that we don't use any personal information since our experiments are based on Carla simulation.
In addition to \textit{\collisionrate} (\collisionrateabbr), \textit{\overallscore} (\overallscoreabbr), and the overall \textit{\selectionrate} (\selectionrateabbr), we also report the \textit{\routecompletion} (\routecompletionabbr) for each scenario before and after selection to measure different algorithms’ ability to influence task performances. 
We find that \AdvTrajabbr achieves the lowest \routecompletionabbr and S-\routecompletionabbr, which demonstrate its effectiveness in attacking the AD system's functionality. 


\subsection{Full benchmark results}
\label{appendix:full_benchmark}

We report the performance of all AD algorithms tested on \name in \Cref{tab:benchmark_full}. 
We trained AD models with different input state spaces and evaluate their performance in both benign scenarios and safety-critical scenarios. 
Specifically, we provide the 4D input to all the $4$ AD algorithms. 
For the 4D+Dir input state, we provide it to SAC, TD3, and PPO. 
We also equip SAC and PPO with both 4D+BEV and 4D+Cam state spaces. 
As shown in the table, we first notice that a large performance gap between evaluation results on benign and safety-critical scenarios always exists no matter what kind of input information we provide to the AD algorithm, which demonstrates that our testing scenarios can generalize to algorithms with different inputs.
Besides, similar to the results of algorithms with 4D input, we also observe the trade-off between performance on benign and safety-critical scenarios in 4D+BEV and 4D+Cam input state spaces. 
For instance, when using 4D+Cam as input state space, SAC obtains a better score on benign scenarios while PPO gets a higher score on safety-critical scenarios.
Finally, among different agents, PPO with 4D+BEV input achieves the best \overallscoreabbr on \name testing scenarios, which indicates potential possible directions for researchers to design their own model architecture and input state space.

\begin{table}[t]
\small
    \centering
    \caption{\small \textbf{The performance of all AD algorithms tested on \name}. We evaluate $4$ algorithms using $4$ different state spaces. We report the average \textit{\overallscore} (\overallscoreabbr) on testing scenarios generated by all the $4$ scenario generation algorithms with driving route variations. \textit{Benign} indicates the performance of AD algorithms tested on normal driving scenarios. The last two columns show the \overallscoreabbr averaged over all benign and safety-critical scenarios. Dir: 4D+Dir, BEV: 4D+BEV, Cam: 4D+Cam.}
    \label{tab:benchmark_full}
    \setlength{\tabcolsep}{3.75pt}
    \begin{tabular}{ll|cccccccc|cc}
    \toprule
    \multirow{2}{*}{\textbf{\makecell{State \\ Space}}} & \multirow{2}{*}{\textbf{Algo.}} & \multicolumn{8}{c|}{\textbf{Traffic Scenarios}} & \multirow{1}{*}{\textbf{Avg}} & \multirow{1}{*}{\textbf{Avg}} \\
    & & \scriptsize{\makecell{Straight \\ Obstacle}} & \scriptsize{\makecell{Turning \\ Obstacle}} & \scriptsize{\makecell{Lane \\ Changing}}  & \scriptsize{\makecell{Vehicle \\ Passing}} & \scriptsize{\makecell{Red-light \\ Running}} & \scriptsize{\makecell{Unprotected \\ Left-turn}} & \scriptsize{\makecell{Right-\\ turn}} & \scriptsize{\makecell{Crossing \\ Negotiation}}
    & \scriptsize{Benign} & \scriptsize{\makecell{Safety-\\critical}} \\
    \midrule
    \multirow{4}{*}{4D} & DDPG & 0.545 & 0.526 & 0.440 & 0.501 & 0.611 & 0.444 & 0.411 & 0.507 & 0.603 & 0.498 \\
    & SAC & 0.533 & 0.474 & 0.577 & 0.471 & 0.482 & 0.501 & 0.503 & 0.432 & \textbf{0.833} & 0.497 \\
    & TD3 & 0.479 & 0.596 & 0.477 & 0.592 & 0.532 & 0.525 & 0.459 & 0.482 & 0.830 & 0.518 \\
    & PPO & 0.761 & 0.611 & 0.426 & 0.432 & 0.755 & 0.728 & 0.605 & 0.655 & 0.819 & \textbf{0.622} \\
    \midrule
    \multirow{3}{*}{Dir} & SAC & 0.608 & 0.591 & 0.670 & 0.435 & 0.624 & 0.548 & 0.552 & 0.522 & 0.752 & 0.569 \\
    & TD3 & 0.728 & 0.543 & 0.499 & 0.451 & 0.665 & 0.595 & 0.645 & 0.590 & \textbf{0.848} & \textbf{0.590} \\
    & PPO & 0.506 & 0.526 & 0.601 & 0.428 & 0.558 & 0.474 & 0.487 & 0.568 & 0.628 & 0.518 \\
    \midrule
    \multirow{2}{*}{BEV} & SAC & 0.501 & 0.567 & 0.647 & 0.446 & 0.486 & 0.521 & 0.449 & 0.434 & \textbf{0.840} & 0.506 \\
    & PPO & 0.818 & 0.632 & 0.555 & 0.393 & 0.918 & 0.664 & 0.729 & 0.847 & 0.731 & \textbf{0.694} \\
    \midrule
    \multirow{2}{*}{Cam} & SAC & 0.634 & 0.570 & 0.436 & 0.427 & 0.481 & 0.529 & 0.527 & 0.425 & \textbf{0.812} & 0.504 \\
    & PPO & 0.542 & 0.503 & 0.407 & 0.425 & 0.928 & 0.519 & 0.579 & 0.808 & 0.613 & \textbf{0.589} \\
    \bottomrule
    \end{tabular}
\end{table}

\subsection{Full diagnostic report}
\label{appendix:full_report}

\begin{table}[t]
\small
    \centering
    \caption{\small \textbf{Diagnostic report of all AD algorithms tested on \name}. We test $4$ AD algorithms with $4$ different state spaces on all selected testing scenarios and report the evaluation results on three different levels. Dir: 4D+Dir, BEV: 4D+BEV, Cam: 4D+Cam.}
    \label{tab:report_full}
    \setlength{\tabcolsep}{3.75pt}
    \begin{tabular}{ll|cccc|ccc|ccc|c}
    \toprule
    \multirow{2}{*}{\textbf{\makecell{State \\ Space}}} & \multirow{2}{*}{\textbf{Algo.}} & \multicolumn{4}{c|}{\textbf{Safety Level}} & \multicolumn{3}{c|}{\textbf{Functionality Level}} & \multicolumn{3}{c|}{\textbf{Etiquette Level}} & \multirow{2}{*}{\textbf{\overallscoreabbr $\uparrow$}} \\
    & & \collisionrateabbr $\downarrow$ & \runredlightabbr $\downarrow$ & \runstopsignabbr $\downarrow$ & \outofroadabbr $\downarrow$ & \followrouteabbr $\uparrow$ & \routecompletionabbr $\uparrow$ & \timespentabbr $\downarrow$ & \accelerationabbr $\downarrow$ & \yawvelocityabbr $\downarrow$ & \laneinvasionabbr $\downarrow$ & \\
    \midrule
    \multirow{4}{*}{4D} & DDPG & 0.780 & 0.089 & 0.087 & 12.619 & 0.504 & 0.466 & 20.860 & 2.488 & 0.405 & 5.764 & 0.489 \\
    & SAC & 0.829 & 0.216 & 0.146 & 3.115 & 0.882 & 0.648 & 16.827 & 1.830 & 0.704 & 2.580 & 0.499 \\
    & TD3 & 0.783 & 0.231 & 0.141 & 2.535 & 0.903 & 0.670 & 17.644 & 2.680 & 1.493 & 2.545 & 0.516 \\
    & PPO & 0.603 & 0.287 & 0.150 & 0.099 & 0.901 & 0.751 & 18.021 & 2.461 & 1.506 & 3.528 & \textbf{0.606} \\
    \midrule
    \multirow{3}{*}{Dir} & SAC & 0.676 & 0.209 & 0.152 & 5.658 & 0.740 & 0.705 & 23.386 & 1.892 & 0.640 & 4.565 & 0.558 \\
    & TD3 & 0.655 & 0.270 & 0.144 & 0.885 & 0.887 & 0.718 & 18.899 & 2.417 & 1.187 & 4.694 & \textbf{0.579} \\
    & PPO & 0.739 & 0.045 & 0.077 & 17.607 & 0.685 & 0.534 & 21.336 & 2.911 & 0.893 & 4.875 & 0.513 \\
    \midrule
    \multirow{2}{*}{BEV} & SAC & 0.782 & 0.229 & 0.141 & 6.057 & 0.883 & 0.674 & 17.863 & 2.952 & 1.566 & 4.448 & 0.506 \\
    & PPO & 0.416 & 0.262 & 0.151 & 2.180 & 0.782 & 0.756 & 30.651 & 2.592 & 1.290 & 7.319 & \textbf{0.679} \\
    \midrule
    \multirow{2}{*}{Cam} & SAC & 0.829 & 0.261 & 0.149 & 0.014 & 0.926 & 0.637 & 15.480 & 4.354 & 1.885 & 6.139 & 0.485 \\
    & PPO & 0.600 & 0.050 & 0.127 & 15.101 & 0.708 & 0.599 & 31.914 & 2.631 & 0.827 & 6.327 & \textbf{0.576} \\
    \bottomrule
    \end{tabular}
\end{table}
\begin{table}[t]
\small
    \centering
    \caption{\small \m{\textbf{Level scores for different levels of evaluation metrics.} We provide $3$ different scores to sum up the safety, functionality, and etiquette levels. The weights are the same as the weights used for the overall score.}}
    \label{tab:report_os_full}
    \setlength{\tabcolsep}{3.75pt}
    \begin{tabular}{ll|ccc|c}
    \toprule
    State Space & Algorithms & Safety & Functionality & Etiquette & Overall \\
    \midrule
    \multirow{4}{*}{4D} & DDPG & 0.459 & 0.541 & 0.755 & 0.489 \\
    & SAC & 0.428 & 0.750 & 0.803 & 0.499 \\
    & TD3 & 0.457 & 0.759 & 0.680 & 0.516 \\
    & PPO & 0.568 & 0.783 & 0.671 & \textbf{0.606} \\
    \midrule
    \multirow{3}{*}{Dir} & SAC & 0.518 & 0.685 & 0.773 & 0.558 \\
    & TD3 & 0.537 & 0.763 & 0.689 & \textbf{0.579} \\
    & PPO & 0.479 & 0.621 & 0.698 & 0.513 \\
    \midrule
    \multirow{2}{*}{BEV} & SAC & 0.450 & 0.753 & 0.629 & 0.506 \\
    & PPO & 0.682 & 0.675 & 0.627 & \textbf{0.679} \\
    \midrule
    \multirow{2}{*}{Cam} & SAC & 0.431 & 0.768 & 0.507 & 0.485 \\
    & PPO & 0.565 & 0.050 & 0.693 & \textbf{0.576} \\
    \bottomrule
    \end{tabular}
\end{table}

In this section, we provide the diagnostic report of all AD algorithms tested on \name. 
We evaluate different combinations of input state spaces and RL algorithms on $3$ different levels of evaluation metrics.
Results are shown in \Cref{tab:report_full}. 
\m{We also provide an overall score for each level in \Cref{tab:report_os_full}.}
We find that PPO achieves the highest \overallscoreabbr in most cases of input, with the highest score of $0.679$ with 4D+BEV state space. 
In addition, regarding the \collisionrate, by comparing agents with different input state spaces, we notice that AD algorithms with 4D input have the highest \collisionrateabbr, while algorithms with 4D+BEV input get the lowest \collisionrateabbr, which indicates that BEV is the most helpful information for AD systems to drive safely.
Finally, we also observe the trade-off between functionality level metrics and safety level metrics with state spaces other than 4D, which means agents that perform well at the functionality level may not be safe regarding the safety level metrics. 
For example, with 4D+BEV input, PPO achieves lower \collisionrateabbr than SAC, while its \routecompletionabbr is also $10.1\%$ lower than SAC. A similar phenomenon can also be found with 4D+Cam input state space.


\newpage

\subsection{Robustness evaluation examples and visualizations}
\label{appendix:toolkit}

\begin{figure}
     \centering
     \includegraphics[width=0.98\textwidth]{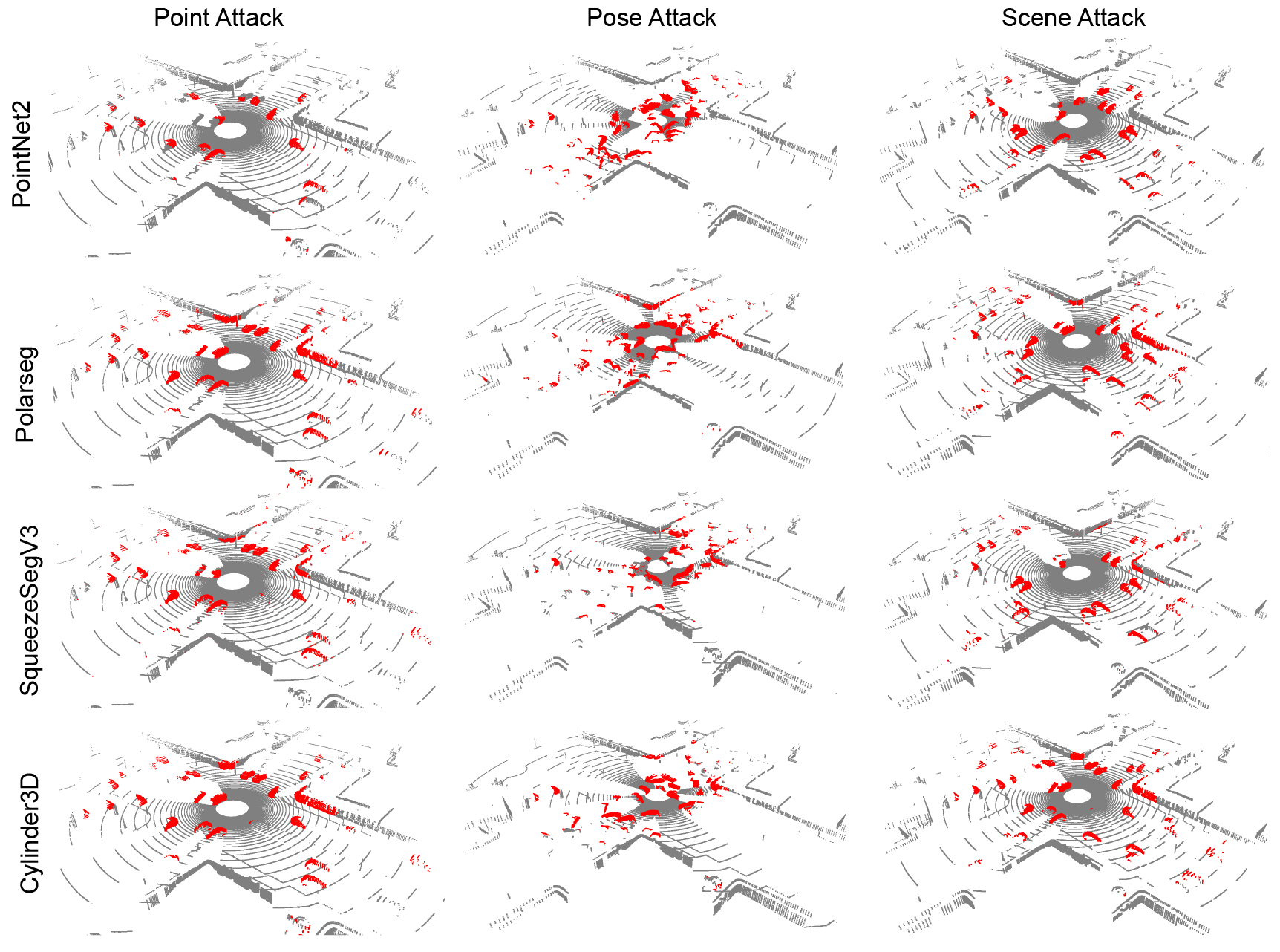}
     \caption{\m{LiDAR point cloud of static traffic scenes generated by $3$ attacking methods (Point Attack, Pose Attack, Scene Attack). Red color means the prediction of $4$ point cloud segmentation algorithms.}}
     \label{fig:lidar_attack}
 \end{figure}

\begin{figure}[t]
     \centering
     \begin{subfigure}[t]{0.24\textwidth}
         \centering
         \includegraphics[width=\textwidth]{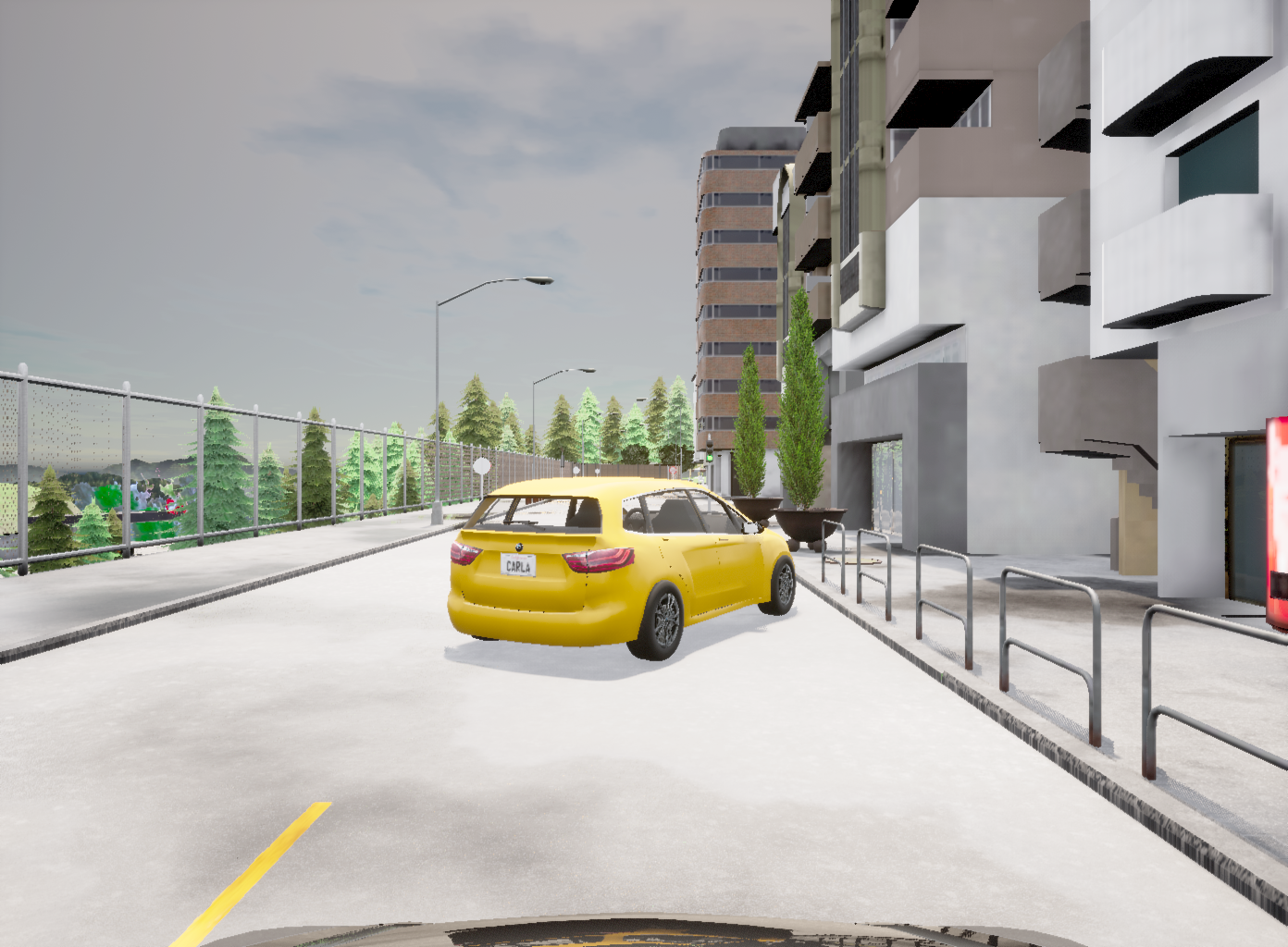}
         \label{subfig:adv1}
     \end{subfigure}
     \hfill
     \begin{subfigure}[t]{0.24\textwidth}
         \centering
         \includegraphics[width=\textwidth]{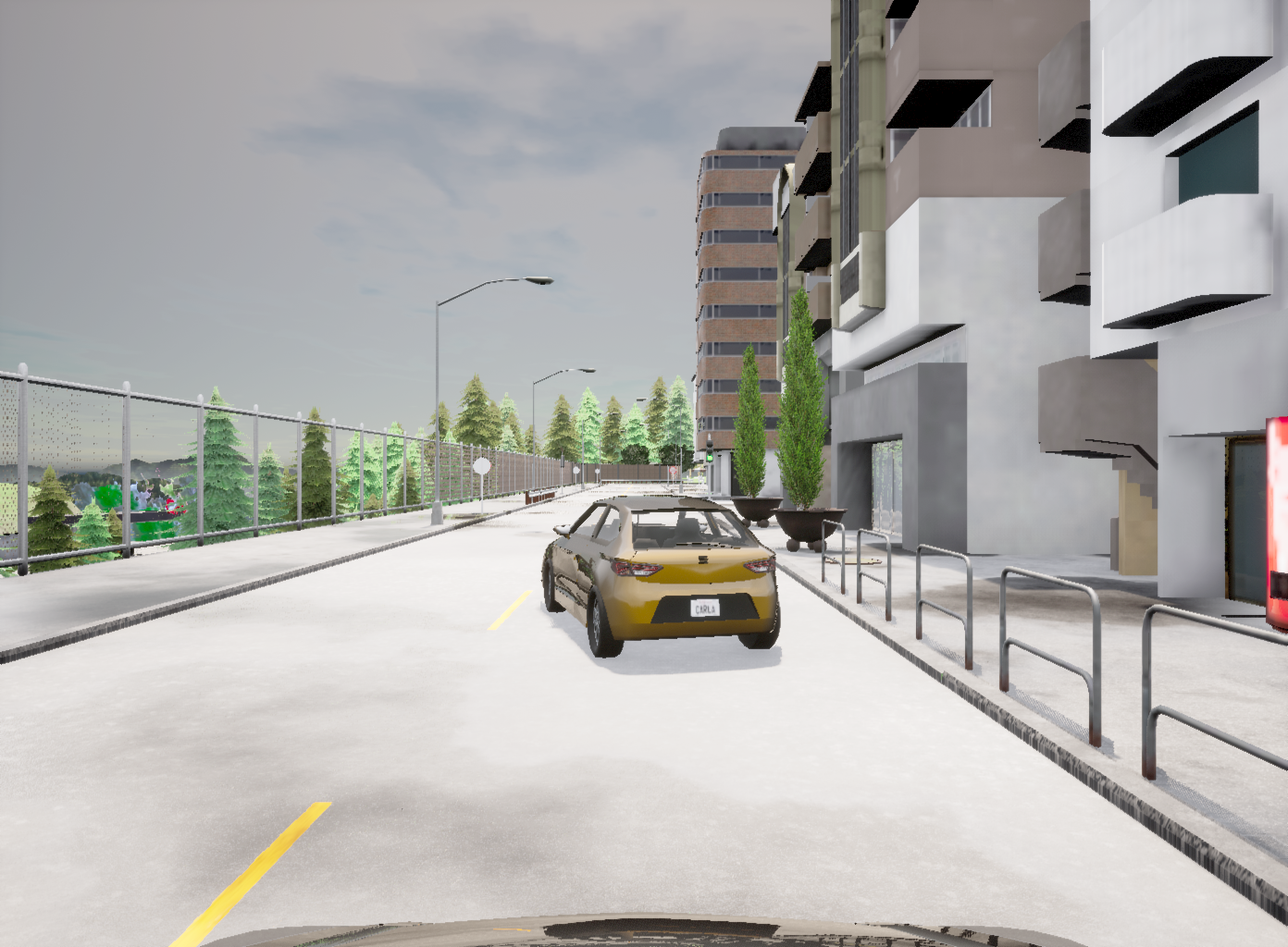}
         \label{subfig:adv2}
     \end{subfigure}
     \hfill
     \begin{subfigure}[t]{0.24\textwidth}
         \centering
         \includegraphics[width=\textwidth]{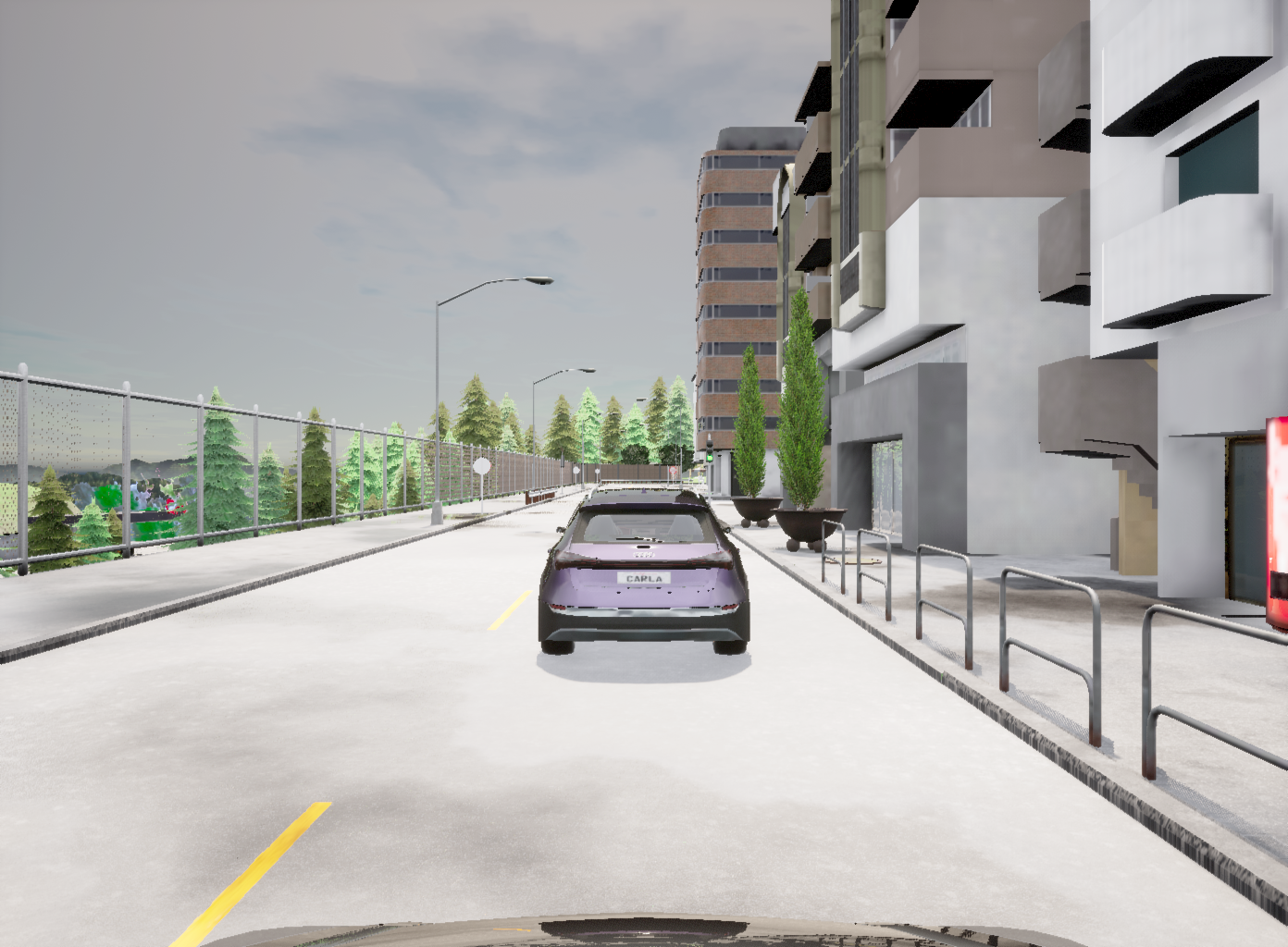}
         \label{subfig:adv3}
     \end{subfigure}
     \hfill
     \begin{subfigure}[t]{0.24\textwidth}
         \centering
         \includegraphics[width=\textwidth]{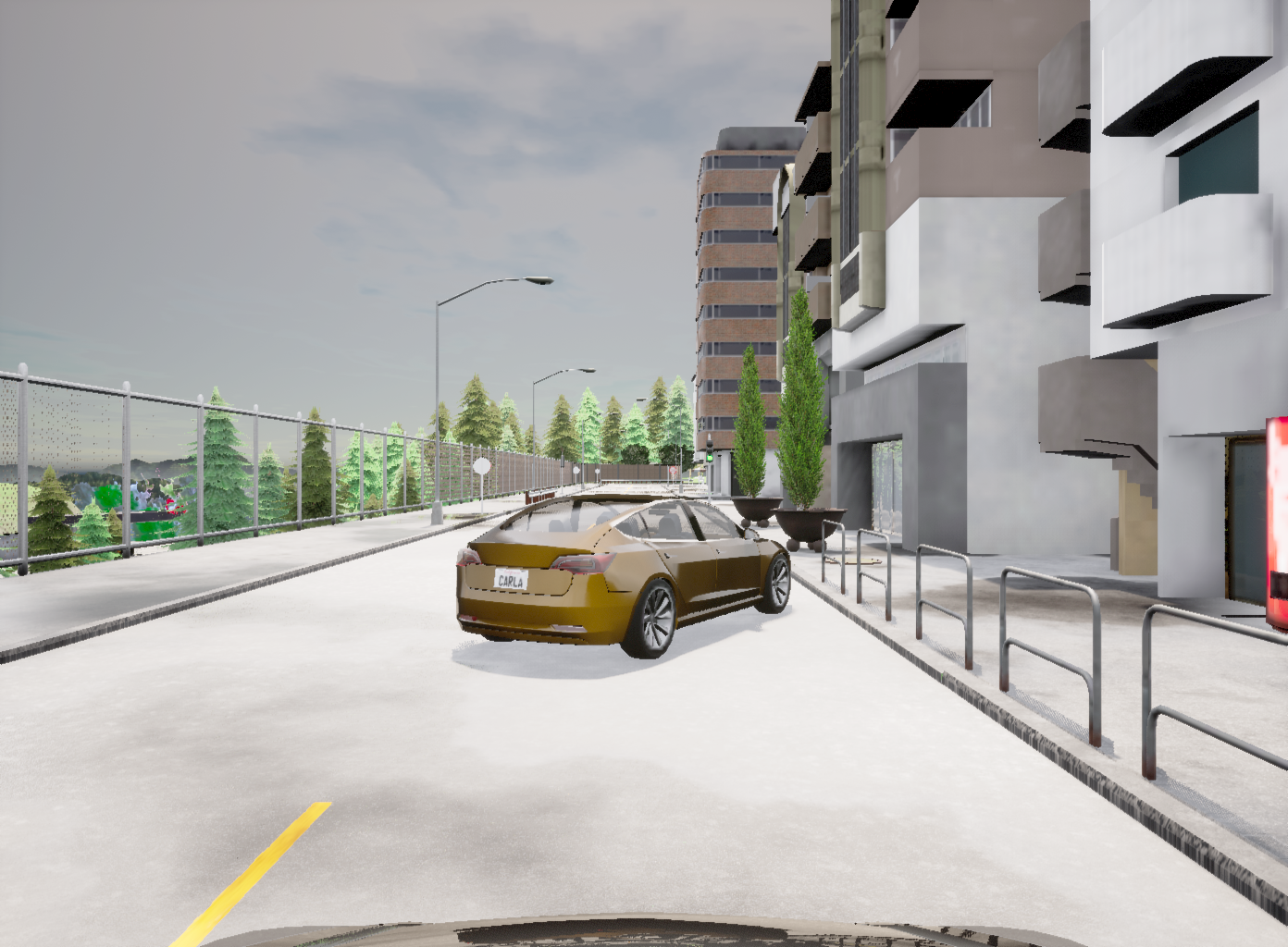}
         \label{subfig:adv4}
     \end{subfigure}
     \hfill
     \begin{subfigure}[t]{0.24\textwidth}
         \centering
         \includegraphics[width=\textwidth]{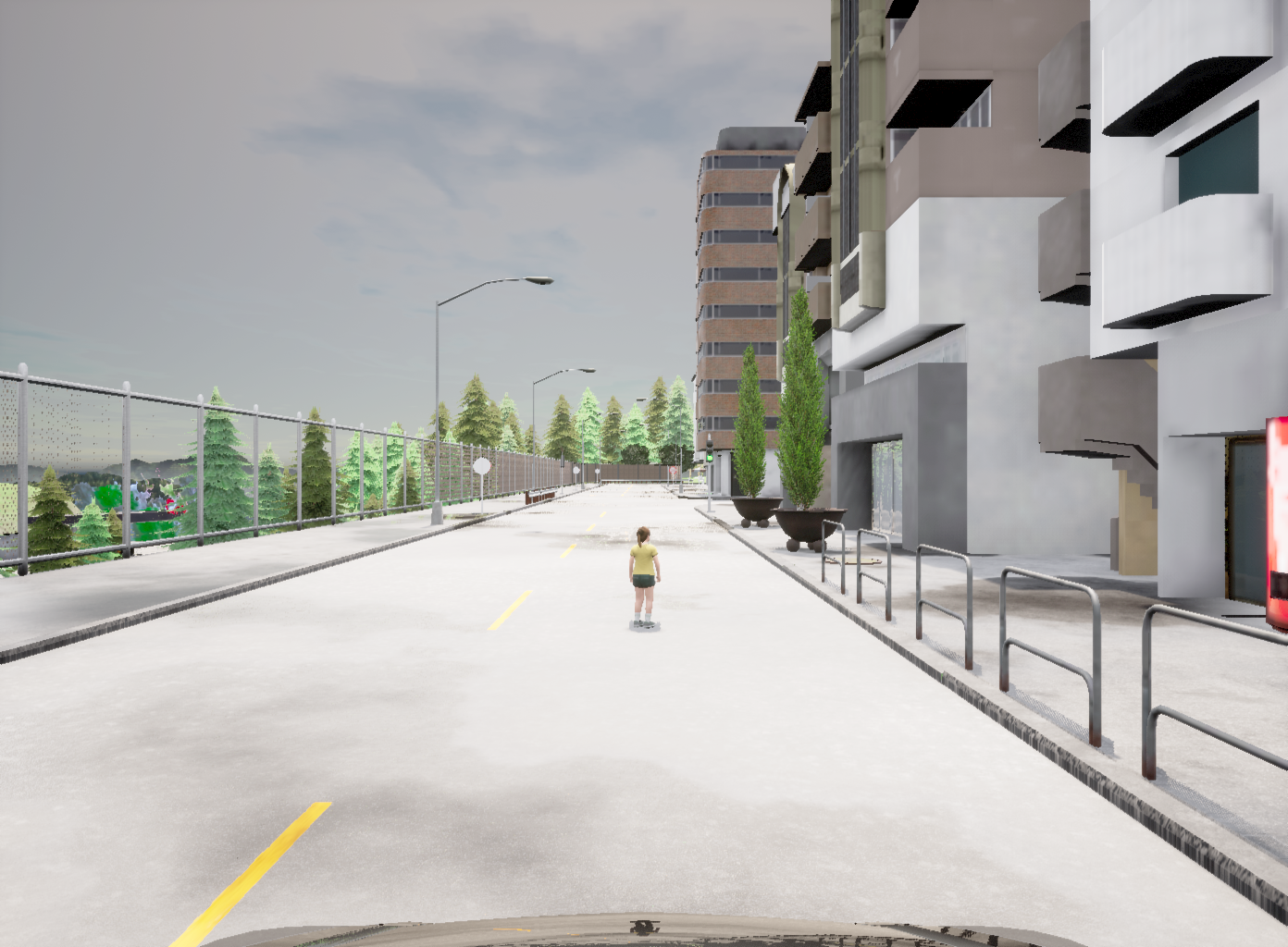}
         \label{subfig:adv5}
     \end{subfigure}
     \hfill
     \begin{subfigure}[t]{0.24\textwidth}
         \centering
         \includegraphics[width=\textwidth]{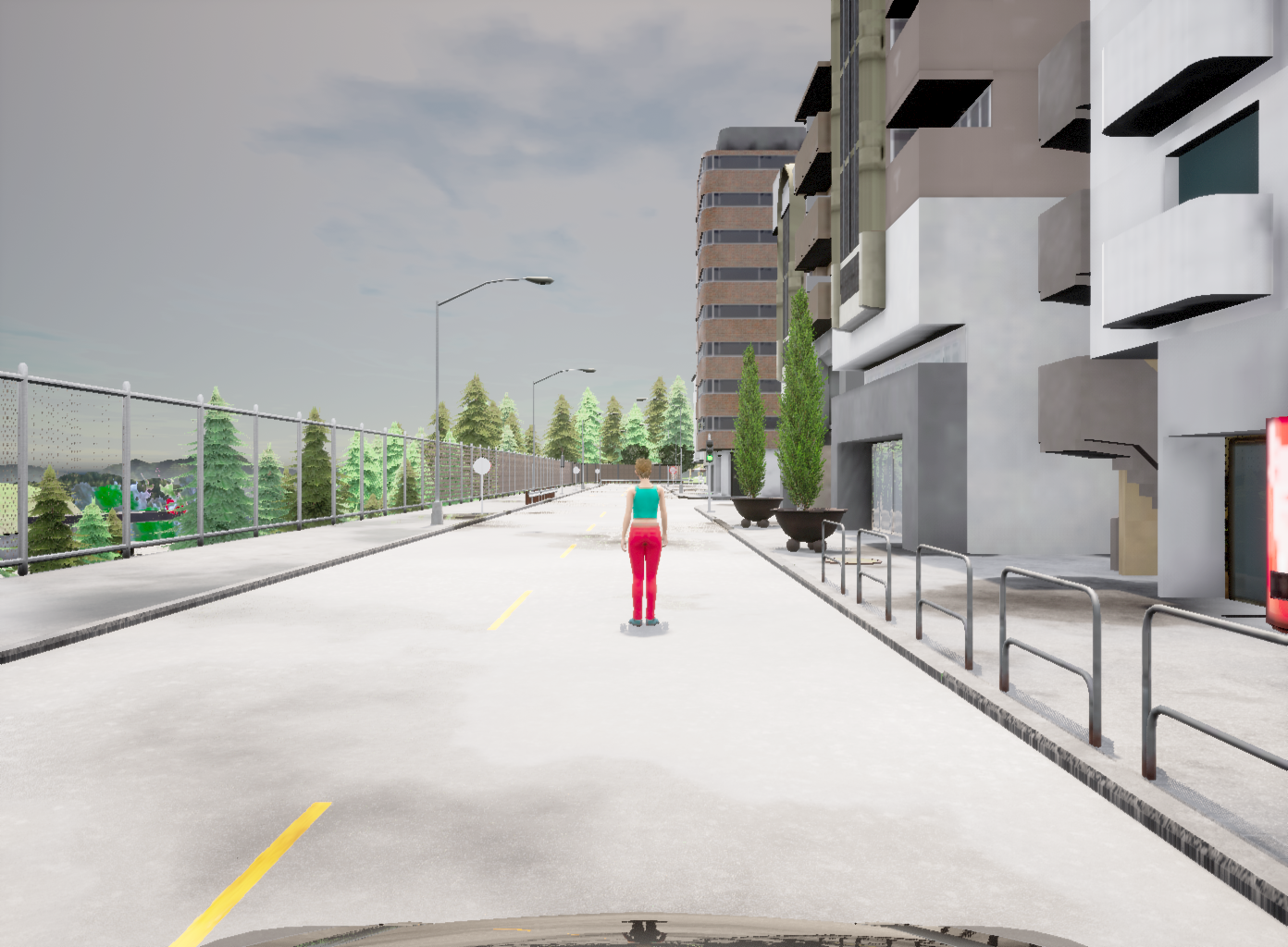}
         \label{subfig:adv6}
     \end{subfigure}
     \hfill
     \begin{subfigure}[t]{0.24\textwidth}
         \centering
         \includegraphics[width=\textwidth]{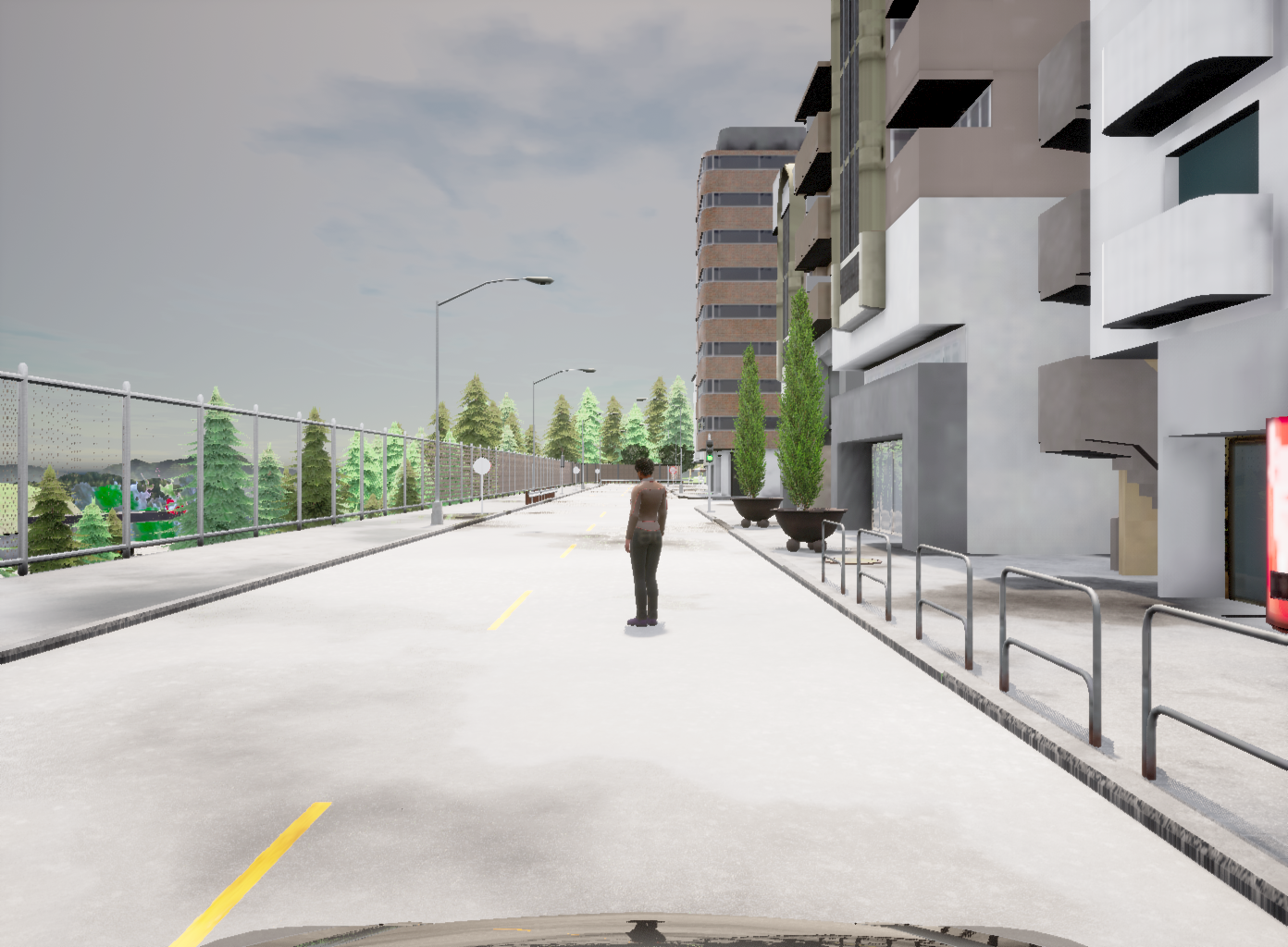}
         \label{subfig:adv7}
     \end{subfigure}
     \hfill
     \begin{subfigure}[t]{0.24\textwidth}
         \centering
         \includegraphics[width=\textwidth]{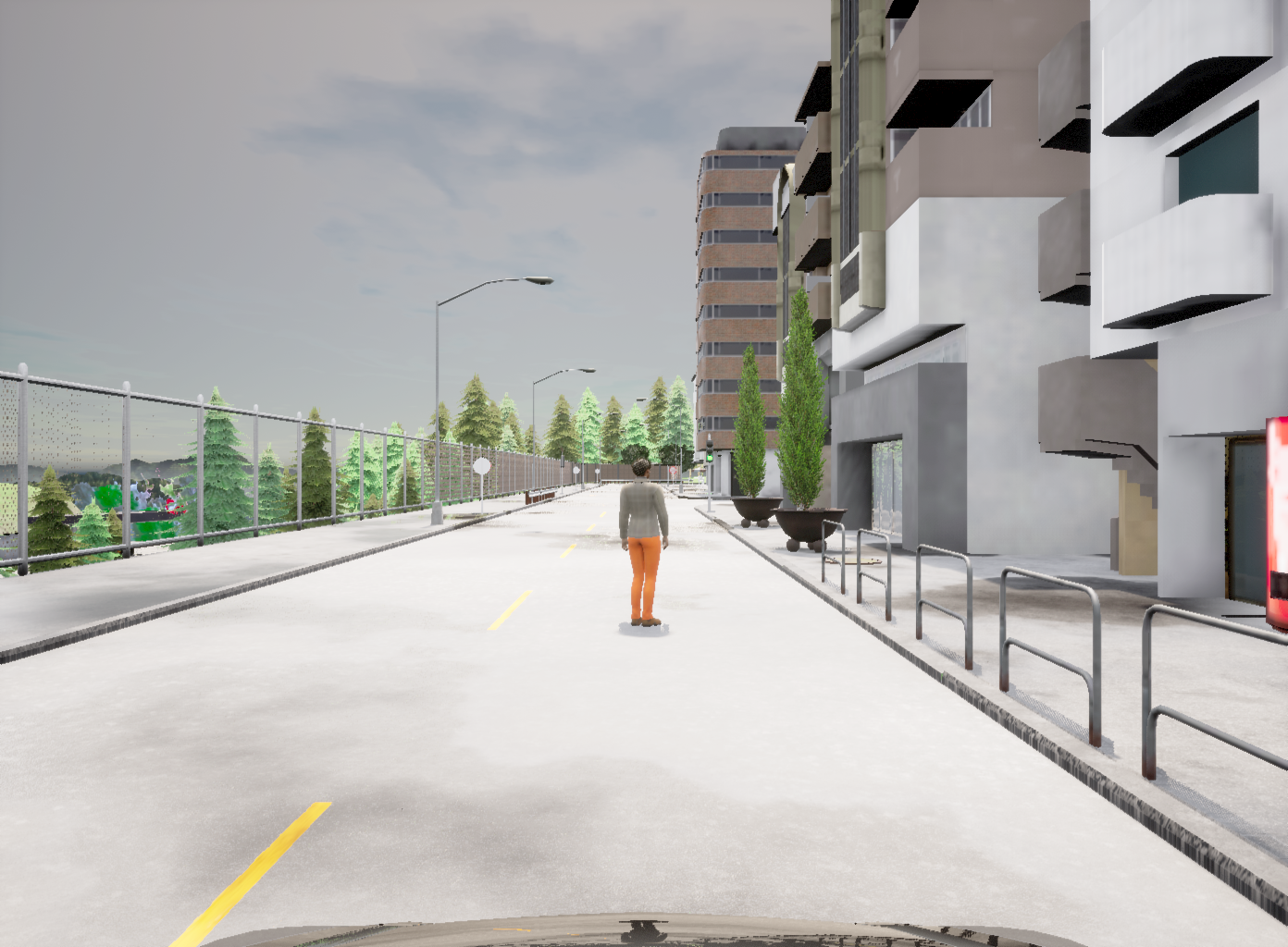}
         \label{subfig:adv8}
     \end{subfigure}
        \caption{\m{Physical semantic perturbations and transformations of vehicles and pedestrians. For vehicles, we consider different types, colors, and rotations. For pedestrians, we consider different body shapes, skin colors, and rotations.}}
        \label{fig:adv}
\end{figure}

\m{In this section, we show detailed examples and visualizations of performing diverse adversarial attacks on AD systems. In~\Cref{fig:lidar_attack}, we provide the adversarial examples of using $3$ different adversarial attacks to attack $4$ different point cloud segmentation models in AD algorithms. In~\cref{fig:adv}, we provide the visualization results of applying $4$ adversarial physical semantic perturbations and transformations to different traffic objects to attack multi-modal object detection models in AD systems.}

\newpage

\subsection{Complexity analysis}

\m{We provide complexity analysis of scenario generation algorithms in $2$ aspects. First, for the algorithm itself, the complexity depends on the searching algorithm inside it. The $4$ scenario generation algorithms considered in \name all use blackbox searching algorithms, but they differ in efficiency. For example, \AdvSim uses Bayesian Optimization, which improves efficiency by prioritizing hyperparameters that appear more promising from past results. Second, for the AD algorithm, the complexity depends on the robustness of the surrogate RL model. A stronger RL algorithm usually needs more optimization steps to reach our safety-critical requirements.}

\subsection{Potential negative societal impacts.}
\label{appendix:impact}
In \name platform, we consider $8$ safety-critical scenarios and design $10$ variations for each scenario.
We also systematically incorporate $4$ scenario generation algorithms with different optimization strategies to fully explore the weakness of AD algorithms.
As we will open-source our platform, attackers may leverage our code and data to perform real-world adversarial attacks against existing AD systems. 
We suggest using our platform to evaluate the safety and robustness of AD systems in various scenarios before deploying them to the real world. 
Since our platform is flexible, developers and researchers can also add more safety-critical scenarios to further test and improve AD systems.

\end{document}